\pdfoutput=1

\documentclass[11pt]{article}

\usepackage[preprint]{acl}

\usepackage[T1]{fontenc}
\usepackage[utf8]{inputenc}
\usepackage{times}
\usepackage{latexsym}
\usepackage{microtype}
\usepackage{inconsolata}

\usepackage{amsmath}
\usepackage{amssymb}
\usepackage{graphicx}

\usepackage{booktabs}
\usepackage{array}
\usepackage{tabularx}
\usepackage{enumitem}
\usepackage[table]{xcolor}

\usepackage[most]{tcolorbox}
\newtcolorbox[auto counter, number within=section]{promptbox}[2][]{
  colback=gray!4,
  colframe=gray!35,
  fonttitle=\small\bfseries,
  title={Prompt~\thetcbcounter: #2},
  label={prompt:#1},
  left=4pt,
  right=4pt,
  top=4pt,
  bottom=4pt,
  breakable
}

\renewcommand{\arraystretch}{1.08}
\newcommand{\lang}[1]{\textbf{\texttt{#1}}}

\title{Prompting Strategy Effects Depend on the Judge: A Multilingual Study of LLM-as-a-Judge Robustness}
\author{
  \textbf{Asal Mehrabi\textsuperscript{1}},
  \textbf{Mohammad Aliannejadi\textsuperscript{1}},
  \textbf{Clemencia Siro\textsuperscript{2}} \\
  \\
  \textsuperscript{1}University of Amsterdam, Amsterdam, The Netherlands \\
  \textsuperscript{2}Centrum Wiskunde \& Informatica (CWI), Amsterdam, The Netherlands \\
  \\
  \small{
    \{\texttt{asalm.1381@gmail.com},
    \texttt{m.aliannejadi@uva.nl},
    \texttt{c.n.siro@cwi.nl}\}
  }
}

\begin{document}
\maketitle

\begin{abstract}
LLM-as-a-Judge evaluation is increasingly applied to multilingual tasks, yet prompting strategies such as Chain-of-Thought and Role Conditioning are recommended based solely on English benchmarks. Whether their effectiveness transfers across judge models and languages remains unknown: prompt-sensitivity studies are largely English-centric, while existing multilingual judge studies fix the prompt template. 
In this work, we address this gap by measuring this interaction through a benchmark of 9 prompting strategies, 4 open-weight judges, 27 languages, and 4 tasks across 13 datasets, yielding over 1M pointwise scores on gold-reference outputs, validated against human judgments. We find that prompting strategies are not transferable: the choice of the judge model dominates the choice of prompting strategy, strategy effects reverse direction across judges, and stability degrades in low-resource languages, where these effects are concealed by aggregate reporting. These findings show that prompting effects in multilingual evaluation are conditioned on the judge, task, and language. Consequently, prompting strategies should be validated for the intended evaluation setting, and cross-study comparisons should account for both judge and prompt configuration.

\end{abstract}

\section{Introduction}
\label{sec:introduction}

Multilingual evaluation of generated text relies on reference-based metrics such as BLEU \citep{papineni-etal-2002-bleu} and ROUGE \citep{lin-2004-rouge}, and on learned metrics such as BLEURT \citep{sellam-etal-2020-bleurt} and COMET \citep{rei-etal-2020-comet}.
These metrics require human-annotated references in the target language.
Such references are unavailable for most languages and costly to produce at scale.
Recent work therefore employs LLMs as evaluators, a paradigm known as LLM-as-a-judge \citep{zheng2023judging, gu2024survey, li2024llmasajudge}.
A prompt template specifies a role, an evaluation rubric, and an output format, and the judge applies the rubric criteria directly rather than scoring against a reference, which extends evaluation to languages without annotated evaluation data.

A reliable multilingual judge should be language-invariant: its scores should reflect the quality of the evaluated output and not the language of that output.
\citet{fu-liu-2025-reliable} score parallel data in 25 languages and report a mean Fleiss' $\kappa$ of approximately $0.3$, with lower agreement on low-resource languages.
Neither multilingual pretraining nor increased model scale improves this consistency.
This instability is distinct from the generation-quality gaps that performance benchmarks report for low-resource languages \citep{ojo-etal-2025-afrobench, li-etal-2025-ssa}: generation capability in a language does not predict evaluation reliability in that language.
The judge model and the prompting strategy remain the configurable components of the evaluation pipeline.

Chain-of-thought \citep{wei2022chain}, few-shot exemplars, self-consistency \citep{wang2023selfconsistency}, and role conditioning \citep{salewski2023} were developed for task solving and later carried over to judging.
Studies of judge prompt sensitivity perturb the paraphrasing or format of one template in English rather than comparing distinct strategies \citep{sclar2024quantifying, bellibatlu2026judgesense}.
\citet{fernandes2023devil} evaluate in-context prompting for MT quality estimation across several PaLM-2 sizes and language pairs, within a single model family.
MM-Eval \citep{son2024mmeval} scales meta-evaluation to more languages and M-Prometheus \citep{pombal2025mprometheus} trains multilingual judges, and neither varies the prompting strategy.
\citet{chung2026crosslingual} show that judge-induced system rankings shift across three related European languages, attributing the instability to the judge's internal language representation rather than to prompt or judge choice, in a setting limited to commercial judges and closely related languages.
Whether these findings extend to open-weight judges, lower-resource languages, and different prompting strategies remains untested. We ask: \textit{Does the effect of a prompting strategy on judge scores depend on the judge model applying it and on the resource level of the language evaluated?}

We benchmark 9 prompting strategies across 4 open-weight judges (Aya Expanse 32B, Gemma 3 27B, Qwen2.5 32B, Qwen2.5 72B), 27 languages, and 4 tasks (machine translation, summarisation, question answering, mathematical reasoning) over 13 datasets, yielding 1.07M pointwise judge scores.
The strategies partition into two groups.
Language-invariant strategies, such as chain-of-thought and few-shot exemplars, hold the prompt language constant.
Language-varying strategies, such as target-language and cross-lingual prompting, vary the prompt language relative to the evaluated content.
The 27 languages span four resource tiers, so resource level varies alongside judge and strategy.
Following \citet{fu-liu-2025-reliable}, we score gold-reference outputs pointwise.
Holding generation quality fixed, attribute score variance to judge configuration rather than to the evaluated content.
A task-specific rubric with a holistic 1--5 rating scores each judge, strategy, and language combination (Section~\ref{sec:methodology}).

Judge identity accounts for $\eta^2 = 0.087$ of score variance, against $\eta^2 = 0.012$ for prompting strategy.
The judge\,$\times$\, strategy interaction reaches $\eta^2 = 0.056$ and exceeds the strategy main effect, indicating that strategy effectiveness is conditional on the judge rather than intrinsic to the prompt.
Target-language prompting raises the mean score of Gemma~3 by $0.83$ and lowers that of Aya Expanse by $0.32$ on identical inputs, with a per-language maximum degradation of $2.16$ on Yoruba.
Inter-judge rank agreement over strategies falls from $\rho = 0.46$ on high-resource machine translation to $\rho = 0.04$ on the low-resource tier, and is negative on summarisation.

Our contributions are as follows.
\begin{enumerate}[label=\textbf{C\arabic*.}, leftmargin=*]
    \item We benchmark prompting-strategy transfer for LLM-as-a-judge evaluation across 9 strategies, 4 judges, 27 languages, 4 tasks, and 13 datasets, and release 1.07M pointwise scores, rubric-level dimension scores, and prompt templates.
    
    \item We decompose score variance across judge, strategy, language, and resource tier, showing that judge identity explains more variance than prompting strategy and that the judge\,$\times$\, strategy interaction exceeds the strategy main effect.
    
    \item We show that target-language prompting produces effects of opposing direction across judges on fixed content and that judge-averaged reporting conceals these differences.
    
    \item We evaluate agreement with human judgements on SSA-MTE, SummEval, and WMT MQM, where Role Prompting attains the highest aggregate correlation ($\rho=0.483$).
\end{enumerate}

\section{Related Work}
\label{sec:related-work}

\paragraph{LLM-as-a-judge.} \citet{zheng2023judging} established that capable LLM judges reach agreement with human annotators comparable to inter-annotator agreement on open-ended generation, motivating LLM-as-a-judge as a substitute for reference-based metrics \citep{papineni-etal-2002-bleu, lin-2004-rouge, sellam-etal-2020-bleurt, rei-etal-2020-comet}. Subsequent work has shifted from validating the paradigm to characterising its failure modes: task complexity, reference quality, and prompt design all affect judge reliability \citep{gu2024survey}, and judge decisions are sensitive to the order in which candidates are presented even when their quality is unchanged \citep{shi2024position}. In multilingual settings, \citet{fu-liu-2025-reliable} evaluate five judges across 25 languages and report a mean Fleiss' $\kappa$ of approximately $0.3$, with agreement deteriorating further for low-resource languages regardless of whether the judge was trained for multilingual use; their prompt-related analysis, however, is confined to a single judge (GPT-4o) and does not compare prompting strategies across architectures. We extend this line by treating judge architecture and prompting strategy as jointly varying factors rather than fixing one to study the other.

\paragraph{Prompt sensitivity and prompting strategies.} Few-shot exemplars \citep{brown2020gpt3}, chain-of-thought reasoning \citep{wei2022chain}, and self-consistency decoding \citep{wang2023selfconsistency} were developed for task-solving and later adapted to judging: rubric-conditioned scoring \citep{kim2024prometheus}, persona conditioning \citep{salewski2023}, and translate-then-evaluate pipelines \citep{fernandes2023devil} are now common judge configurations, alongside calibration-example prompting for translation quality estimation \citep{kocmi2023gemba}. \citet{bellibatlu2026judgesense} introduce the Judge Sensitivity Score to quantify how often a judge changes its decision under semantically equivalent rephrasing, evaluating 13 judges across four tasks and finding that model scale does not predict stability; their study, like the prompting literature it builds on, is English-only, leaving open whether prompt sensitivity itself is language-dependent. \citet{sclar2024quantifying} report a related finding for task-solving LLMs, that formatting perturbations of a fixed prompt shift performance substantially, again in English. We adapt the comparative logic of these studies, holding content fixed and varying the prompt to a multilingual, multi-judge setting.

\paragraph{Multilingual and low-resource NLP.} Benchmark infrastructure for low-resource languages has expanded through Masakhane \citep{masakhane2021}, FLORES-200 \citep{nllb2024nature}, and MAFAND-MT \citep{adelani2022mafand}, yet models continue to underperform on typologically diverse and low-resource languages independent of training-data volume \citep{lowresourceindiclanguages2025}, and cross-lingual transfer remains contingent on linguistic similarity \citep{qiu2024crosslingualrransfer}. Within LLM-as-a-judge specifically, \citet{chung2026crosslingual} generate synthetic dialogues under matched conditions across three Finno-Ugric languages and find that surface-level dimensions (grammar, fluency) are stable across languages while higher-level judgments (coherence) are not, indicating that some cross-lingual instability originates in the evaluation process itself rather than in the content evaluated. \citet{chang2025multilingual} show that judge--human alignment declines as resource level falls, consistent with \citeauthor{fu-liu-2025-reliable}'s findings. MM-Eval \citep{son2024mmeval} and M-Prometheus \citep{pombal2025mprometheus} extend meta-evaluation and judge training to more languages, respectively, but both fix the prompting protocol; neither varies it as a factor. No existing study crosses prompting strategy with judge architecture and language resource level simultaneously, which is the gap we address.

\section{Methodology}
\label{sec:methodology}

\subsection{Experimental Design}
\label{sec:design}

We examine the effects of prompting strategy and judge model on multilingual pointwise evaluation. The benchmark comprises four open-weight judges, nine prompting strategies, 27 languages, four tasks, and 13 datasets. Each observation is a score assigned to a candidate output by one judge under one prompting strategy.

All applicable judge--strategy configurations receive the same source input, candidate output, and task-specific rubric. Except for SSA-MTE, the candidate outputs are human-written references. We separately evaluate alignment with human quality judgements using system outputs from SSA-MTE, SummEval, and WMT MQM.

\subsection{Tasks and Datasets}
\label{sec:datasets}

We evaluate machine translation (MT), summarisation, question answering (QA), and mathematical reasoning. Table~\ref{tab:datasets} reports the evaluated language splits and sample sizes. Unless otherwise specified, we randomly sample up to 200 instances per dataset--language split using seed 42.

\begin{table*}[t]
\centering
\small
\setlength{\tabcolsep}{5pt}
\renewcommand{\arraystretch}{0.95}
\begin{tabular}{llrrr}
\toprule
\textbf{Task} & \textbf{Dataset} & \textbf{\#Lang.} &
\textbf{Instances/split} & \textbf{Total} \\
\midrule
MT
& FLORES-200~\citep{nllb2024nature}
& 14 & $\leq$200 & 2,395 \\
MT
& MAFAND-MT~\citep{adelani2022mafand}
& 10 & $\leq$200 & 1,998 \\
MT
& SSA-MTE~\citep{ojo2025ssamte}
& 11 & 848--1,503 & 12,565 \\
\midrule
Summarisation
& XL-Sum~\citep{hasan2021xlsum}
& 13 & $\leq$200 & 2,598 \\
Summarisation
& MLSUM~\citep{scialom2020mlsum}
& 2 & 200 & 400 \\
Summarisation
& CaSum~\citep{casum2022}
& 1 & 200 & 200 \\
\midrule
QA
& AfriQA~\citep{ogundepo2023afriqa}
& 9 & 196--200 & 1,795 \\
QA
& XQuAD~\citep{artetxe2020xquad}
& 3 & 200 & 600 \\
QA
& PersianQA~\citep{ayoubi2021persianqa}
& 1 & All & 140 \\
QA
& CatalanQA~\citep{armengol2022catalanqa}
& 1 & 200 & 200 \\
QA
& Belebele~\citep{bandarkar2024belebele}
& 14 & $\leq$900 & 12,598 \\
\midrule
Mathematics
& AfriMGSM~\citep{adelani2024afrimgsm}
& 16 & 8 & 128 \\
Mathematics
& MGSM~\citep{shi2022mgsm}
& 4 & 8 & 32 \\
\bottomrule
\end{tabular}
\caption{Evaluation datasets and sample sizes. ``All'' indicates that all available instances were retained.}
\label{tab:datasets}
\end{table*}

MT instances contain a source sentence and candidate translation. Summarisation instances contain a source document and candidate summary. QA instances contain a question and candidate answer. XQuAD, PersianQA, CatalanQA, and Belebele additionally provide a supporting passage. Mathematical-reasoning instances contain a problem and an annotated solution comprising intermediate reasoning and a final answer.

\paragraph{Language resource groups.}
We partition the 27 languages into four resource groups:
\textit{high} ($n{=}4$), comprising \lang{En} (English), \lang{Fr}
(French), \lang{De} (German), and \lang{Es} (Spanish);
\textit{mid} ($n{=}3$), comprising \lang{Ca} (Catalan), \lang{Fa}
(Persian), and \lang{Ar} (Arabic);
\textit{low--mid} ($n{=}2$), comprising \lang{Hau} (Hausa) and
\lang{Swa} (Swahili); and
\textit{low} ($n{=}18$), comprising \lang{Yor} (Yoruba), \lang{Ibo}
(Igbo), \lang{Kin} (Kinyarwanda), \lang{Xho} (Xhosa), \lang{Zul}
(Zulu), \lang{Pcm} (Nigerian Pidgin), \lang{Twi} (Twi), \lang{Bem}
(Bemba), \lang{Fon} (Fon), \lang{Am} (Amharic), \lang{Sot} (Sotho),
\lang{Tir} (Tigrinya), \lang{Ewe} (Ewe), \lang{Lug} (Luganda),
\lang{Lin} (Lingala), \lang{Orm} (Oromo), \lang{Som} (Somali), and
\lang{Wol} (Wolof).
Hausa and Swahili are the only languages represented across all four
tasks. Appendix~\ref{app:language-coverage} reports the language coverage
of each dataset.

\subsection{Prompting Strategies}
\label{sec:strategies}

We evaluate nine prompting strategies spanning calibration examples, explicit reasoning, evaluator roles, and prompt language. All strategies use the same task-specific rubric and five-point scoring scale. Table~\ref{tab:strategies} reports their task applicability.

\begin{table}[t]
\centering
\small
\setlength{\tabcolsep}{3.2pt}
\renewcommand{\arraystretch}{0.95}
\begin{tabular}{lcccc}
\toprule
\textbf{Strategy} & \textbf{MT} & \textbf{Sum.} &
\textbf{QA} & \textbf{Math} \\
\midrule
Zero-Shot         & \checkmark & \checkmark & \checkmark & \checkmark \\
Few-Shot Anchored & \checkmark & \checkmark & \checkmark & \checkmark \\
Prometheus-style  & \checkmark & \checkmark & \checkmark & \checkmark \\
Chain-of-Thought  & --         & \checkmark & \checkmark & \checkmark \\
Role Prompting    & \checkmark & \checkmark & \checkmark & -- \\
Self-Consistency  & --         & --         & \checkmark & \checkmark \\
Cross-Lingual     & \checkmark & \checkmark & \checkmark & \checkmark \\
Monolingual       & \checkmark & \checkmark & \checkmark & \checkmark \\
Trans-Eval        & \checkmark & \checkmark & --         & -- \\
\bottomrule
\end{tabular}
\caption{Prompting strategies and task applicability.}
\label{tab:strategies}
\end{table}

\paragraph{Reasoning and calibration strategies.}
Zero-shot provides the evaluation instruction, rubric, score definitions, and instance without demonstrations. Few-shot anchored adds high- and low-quality calibration examples~\citep{fernandes2023devil,kocmi2023gemba}. Prometheus-style prompting elicits criterion-level assessments before the holistic score~\citep{kim2024prometheus}. Chain-of-Thought requests an explicit stepwise assessment before scoring~\citep{wei2022chain}. Role prompting assigns a task- and language-specific evaluator role~\citep{salewski2023}. Self-consistency samples three Chain-of-Thought evaluations and aggregates their holistic scores~\citep{wang2023selfconsistency}.

\paragraph{Language-handling strategies.}
Cross-Lingual prompting uses English instructions to evaluate target-language candidates~\citep{ahuja2023mega}. Monolingual prompting translates the instructions, rubric, and score definitions into the candidate language. Trans-Eval first translates the candidate into English and then evaluates the translated output~\citep{fernandes2023devil}. Appendix~\ref{app:prompts} provides the complete prompt templates.

\subsection{Evaluation Protocol}
\label{sec:evaluation}

Each judge assigns task-specific criterion scores and a holistic score from 1 to 5, with higher scores indicating better quality. We use the holistic score for cross-task analyses and the criterion scores for task-specific analyses. Table~\ref{tab:rubrics} summarises the criteria.

\begin{table}[t]
\centering
\small
\setlength{\tabcolsep}{3.5pt}
\begin{tabular}{@{}p{0.9cm}p{3.45cm}p{1.9cm}@{}}
\toprule
\textbf{Task} & \textbf{Criteria} & \textbf{Basis} \\
\midrule
MT
& Adequacy, fluency, terminology
& MQM~\citep{freitag2021experts} \\
Sum.
& Comprehensibility, attribution, main-idea coverage, conciseness
& SEAHORSE
~\citep{clark2023seahorse}; SummEval~\citep{fabbri2021summeval} \\
QA
& Relevance, completeness, passage grounding
& MM-Eval~\citep{son2024mmeval}; Prometheus~\citep{kim2024prometheus} \\
Math
& Answer correctness, reasoning validity, solution completeness
& Process Reward~\citep{lightman2023verify}; GSM8K~\citep{cobbe2021gsm8k} \\
\bottomrule
\end{tabular}
\caption{Task-specific evaluation criteria.}
\label{tab:rubrics}
\end{table}

For XQuAD, PersianQA, CatalanQA, and Belebele, passage grounding measures whether the answer is supported by the supplied passage. AfriQA provides no passage, so its rubric contains only relevance and completeness. For mathematical reasoning, judges assess the final answer, validity of the derivation, and completeness of the solution.

We evaluate human-score alignment on SSA-MTE~\citep{ojo2025ssamte}, SummEval~\citep{fabbri2021summeval}, and WMT MQM~\citep{freitag2021experts}. SSA-MTE contains 12,565 multilingual system translations, SummEval contains 1,600 English summaries, and the WMT MQM subset contains 500 English translations.

\subsection{Judge Models and Implementation Details}
\label{sec:implementation}

We use four instruction-tuned open-weight judges: Aya Expanse 32B~\citep{ustun2024aya}, Gemma 3 27B~\citep{gemma3_2025}, Qwen 2.5 32B, and Qwen 2.5 72B-AWQ~\citep{qwen2024qwen25}. 

We run batched inference with vLLM~\citep{kwon2023vllm} on NVIDIA A100 and H100 GPUs. All strategies except self-consistency use temperature 0 and top-$p=1.0$. Self-Consistency uses temperature 0.7 and three independently sampled responses. We select the modal holistic score. When all three scores differ, we use their arithmetic mean. We use seed 42 throughout.

Each response contains criterion-level scores, a holistic score, and a short justification. We extract numeric scores from the specified response schema. Outputs without a valid holistic score in the interval $[1,5]$ are recorded as parsing failures and excluded from score aggregation. The evaluation produced 1,068,528 valid scores.

\subsection{Statistical Analysis}
\label{sec:analysis}

We compute mean holistic scores within judge--strategy--task--language cells. The range of language-level means quantifies absolute cross-language variation. We quantify relative cross-language dispersion using the coefficient of variation,
\[
\mathrm{CV}_{j,p,t}
=
\frac{\sigma_{j,p,t}}{\mu_{j,p,t}},
\]
where $\mu_{j,p,t}$ and $\sigma_{j,p,t}$ denote the mean and standard deviation across languages for judge $j$, strategy $p$, and task $t$. We compute strategy comparisons and CV separately by task.

We measure inter-judge agreement using Spearman's $\rho$ for strategy rankings and Pearson's $r$ for strategy-level mean scores. We estimate the following two-way ANOVA separately for each task:
\[
s
=
\mathrm{Judge}
+
\mathrm{Strategy}
+
\mathrm{Judge}\times\mathrm{Strategy}.
\]
Separate one-way ANOVAs quantify variation associated with language and resource group. We report Type-II eta squared,
\[
\eta_f^2
=
\frac{\mathrm{SS}_f}{\mathrm{SS}_{\mathrm{total}}},
\]
where $\mathrm{SS}_f$ is the sum of squares associated with factor $f$. We assess ANOVA terms at $\alpha=0.001$. We report pairwise strategy comparisons descriptively rather than significance-testing them, because ceiling effects in QA and reference-based MT compress the score range and would inflate rejection rates.

For each task and strategy, we compute the difference between the English mean and the low-resource-language mean. We measure the language-specific effect of Monolingual prompting relative to Zero-Shot as
\[
\Delta_{\mathrm{mono}}(l,j,t)
=
\bar{s}_{\mathrm{Monolingual},l,j,t}
-
\bar{s}_{\mathrm{Zero\mbox{-}Shot},l,j,t}.
\]
We exclude English because both conditions use English instructions. Languages represented in only one task are excluded from cross-task comparisons.

Finally, we compute instance-level Spearman correlations between holistic judge scores and human annotations on SSA-MTE ($n \approx 12{,}579$ per cell), SummEval ($n = 1{,}600$), and WMT MQM ($n = 500$). We estimate correlations separately for each judge and applicable prompting strategy.

\section{Results}
\label{sec:results}

We decompose score variance across judge, strategy, and their interaction (\S\ref{sec:rq1}), measure how strategy effects vary across judges (\S\ref{sec:rq2}) and languages (\S\ref{sec:rq3}), and report agreement with human annotations (\S\ref{sec:validity}).

Scoring gold-reference outputs yields mean scores above $4.5$ on QA and reference-based MT, compressing the range over which strategies can differ. A supplementary evaluation on AfroBench system outputs \citep{ojo-etal-2025-afrobench} yields wider dispersion than the gold-reference condition for the same judges (Appendix~\ref{app:ceiling}). Gemma~27B assigns lower Zero-Shot scores than the other judges ($2.29$ QA, $3.15$ Summarization, $2.90$ Math), and its scores under the other strategies fall within the range of the other judges (Appendix~\ref{app:heatmaps}).

\subsection{Variance Decomposition}
\label{sec:rq1}

Table~\ref{tab:eta_squared} reports $\eta^2_f$ from a two-way ANOVA over judge and strategy, with language and resource estimated by separate one-way ANOVAs. All terms are significant at $p<0.001$.

Judge identity accounts for $\eta^2=0.087$ of score variance and prompting strategy for $\eta^2=0.012$, a ratio of approximately seven to one. The judge $\times$ strategy interaction is $\eta^2=0.056$ and exceeds the strategy main effect.

\paragraph{Per task.} Judge identity accounts for $\eta^2=0.438$ on Summarization. On QA the interaction is $\eta^2=0.247$ against $0.251$ for judge identity, and on Math $0.297$ against $0.222$. On MT, language accounts for $\eta^2=0.325$ and judge identity for $\eta^2=0.006$. Excluding SSA-MTE, the only MT dataset containing system outputs with cross-lingual quality variation, gives a language term of $0.030$, a judge term of $0.048$, and an interaction of $0.114$ (Appendix~\ref{app:eta}).

We also observe that Gemma~27B scores substantially lower under Zero-Shot than the other judges but recovers under other strategies, contributing to its distinctive per-task profile (Appendix~\ref{app:heatmaps}).

\begin{table}[!t]
\centering
\small
\setlength{\tabcolsep}{3pt}
\begin{tabular}{@{}lrrrrr@{}}
\toprule
\textbf{Factor} & \textbf{All} & \textbf{MT} &
\textbf{Sum.} & \textbf{QA} & \textbf{Math} \\
\midrule
Judge
  & 0.087 & 0.006 & 0.438 & 0.251 & 0.222 \\
Strategy
  & 0.012 & 0.007 & 0.057 & 0.059 & 0.036 \\
Judge $\times$ strategy
  & 0.056 & 0.017 & 0.165 & 0.247 & 0.297 \\
Language$^\dagger$
  & 0.043 & 0.325 & 0.032 & 0.012 & 0.010 \\
Resource$^\dagger$
  & 0.007 & 0.051 & 0.010 & 0.003 & 0.001 \\
\bottomrule
\end{tabular}
\caption{Variance decomposition ($\eta^2$). All effects have
$p<0.001$. $^\dagger$One-way ANOVA. Results for MT excluding SSA-MTE are
reported in Appendix~\ref{app:eta}.}
\label{tab:eta_squared}
\end{table}

\subsection{Strategy Effects Across Judges}
\label{sec:rq2}

Table~\ref{tab:rank_agreement} reports mean Spearman $\rho$ over the six judge pairs. On MT, agreement is $\rho=0.46$ on the high-resource tier and $\rho=0.04$ on the low-resource tier. Summarization yields negative values in three of four tiers ($-0.30$ to $-0.02$). QA values lie between $-0.08$ and $+0.02$.

Pearson correlations on strategy-level means differ by judge pair. The two Qwen judges correlate at $r=0.819$. Aya correlates with Q32 at $r=0.385$ and with Q72 at $r=0.411$. Gemma correlates with the other three between $r=-0.173$ and $r=0.025$. Full matrices are in Appendix~\ref{app:intermodel_full}.

\begin{table}[htbp]
\centering
\small
\setlength{\tabcolsep}{5pt}
\begin{tabular}{l rrrr}
\toprule
\textbf{Task} & \textbf{High} & \textbf{Mid} & \textbf{LM} & \textbf{Low} \\
\midrule
MT  & $+$0.46 & $+$0.02 & $+$0.39 & $+$0.04 \\
Sum & $-$0.30 & $-$0.20 & $+$0.10 & $-$0.02 \\
QA  & $-$0.00 & $-$0.08 & $+$0.02 & $-$0.04 \\
\bottomrule
\end{tabular}
\caption{Inter-judge rank agreement (mean Spearman $\rho$ over 6 judge pairs) by task and resource tier. LM = low-mid. Math excluded for insufficient cross-tier coverage.}
\label{tab:rank_agreement}
\end{table}

\subsection{Strategy Effects Across Languages}
\label{sec:rq3}

\begin{figure}[t]
  \centering
  \includegraphics[width=1\columnwidth]{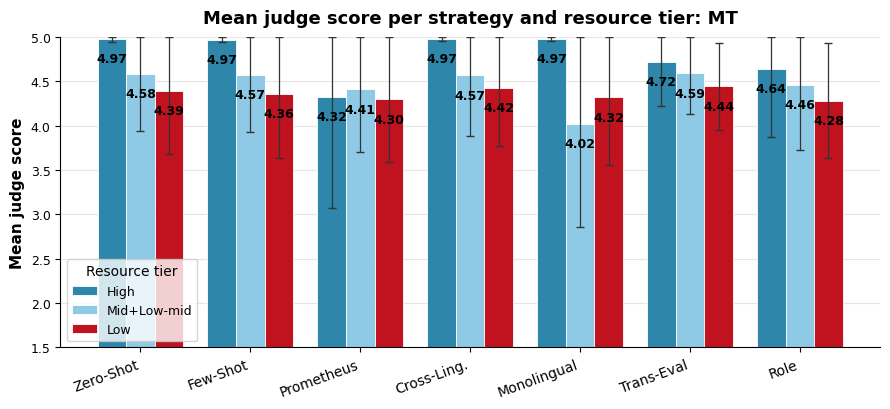}\\[2pt]
  \includegraphics[width=1\columnwidth]{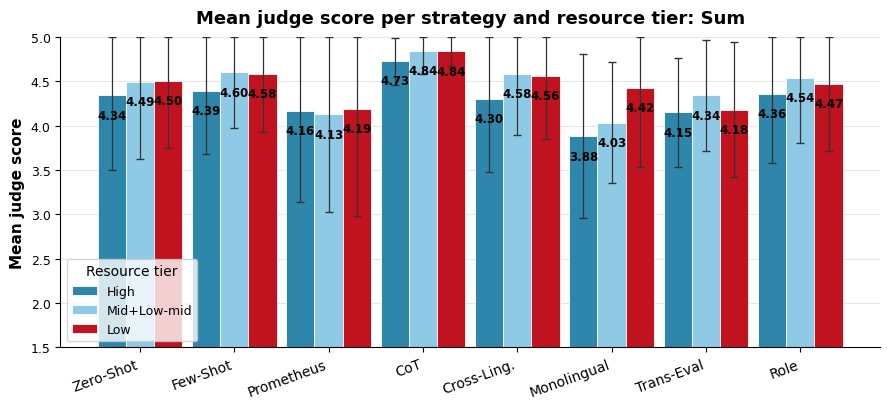}
  \caption{Mean judge score per strategy and resource tier for MT
  (top) and Summarization (bottom), averaged over 4 judges.
  QA and Math panels in Appendix~\ref{app:heatmaps}.}
  \label{fig:rq1_overview}
\end{figure}

Figure~\ref{fig:rq1_overview} reports mean judge score per strategy and resource tier for MT and Summarization. CoT has the highest mean on Summarization, QA, and Math (Appendix~\ref{app:heatmaps}). Monolingual has the lowest mean on MT and Summarization, and Prometheus the lowest on QA and Math.

Cross-lingual ranges are wider at lower resource levels (Table~\ref{tab:rq1_range}). On MT the mean range is $0.13$ on the high-resource tier and $1.60$ on the low-resource tier. At the rubric level the difference is largest on Terminology, which is $3.93$--$4.59$ on low-resource languages against $4.97$--$5.00$ on high-resource languages, while Adequacy and Fluency differ by at most $0.5$ points across tiers (Appendix~\ref{app:rubric_dims}).

Mean $\mathrm{CV}$ ranges from $0.036$ for CoT to $0.165$ for Prometheus, with Monolingual second highest at $0.162$ (Appendix~\ref{app:cv_full}). Under CoT and Prometheus, the Summarization Main Ideas dimension is $2.99$--$3.07$ against $3.48$--$3.80$ under Zero-Shot, across all four resource tiers.

\begin{figure}[!t]
  \centering
  \includegraphics[width=\columnwidth]{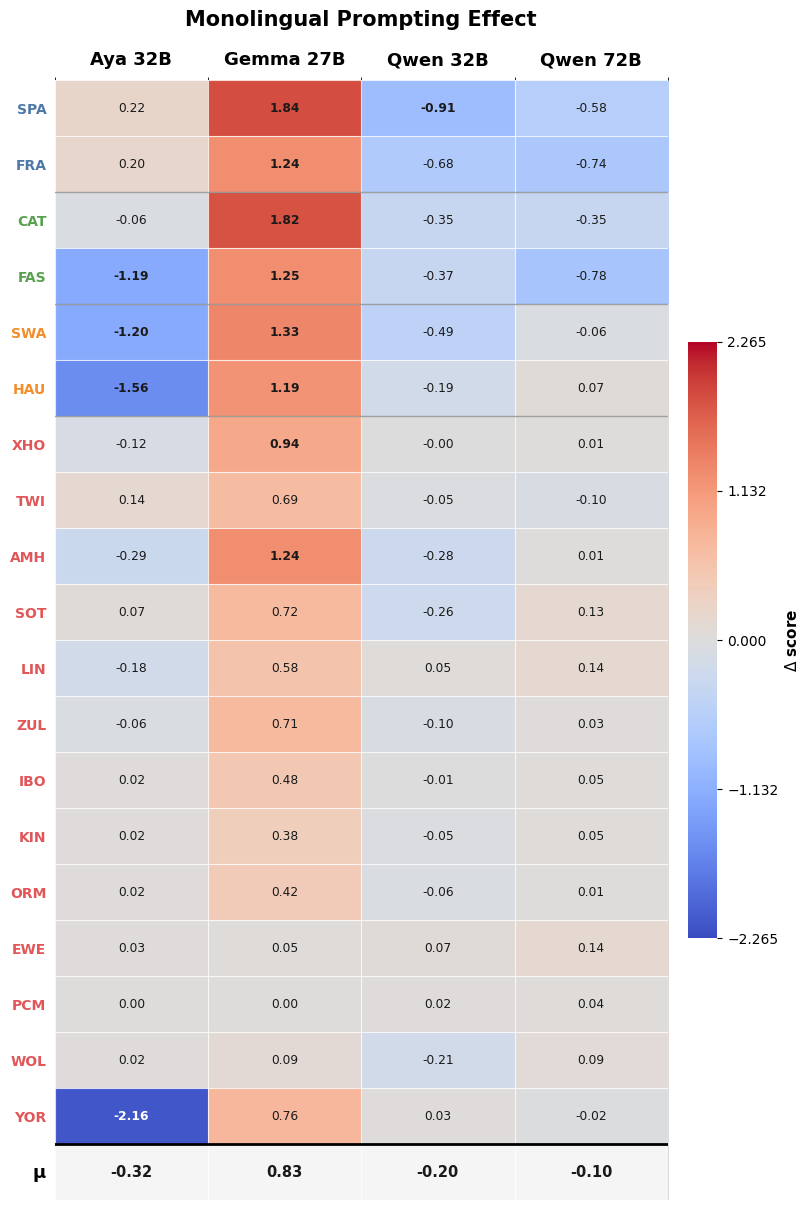}
  \caption{Monolingual prompting effect by language and judge
  ($\Delta$ = Monolingual $-$ Zero-Shot). Rows ordered by resource
  tier, then mean effect; bottom row ($\mu$) = mean per judge.}
  \label{fig:rq3_mono}
\end{figure}

Figure~\ref{fig:rq3_mono} reports $\Delta_{\mathrm{mono}}$ per language and judge. Monolingual prompting raises Gemma's mean score by $0.83$ and lowers Aya's by $0.32$; the two Qwen judges shift by $-0.20$ and $-0.10$. Aya's largest per-language decreases are on \lang{Yor} ($-2.16$), \lang{Hau} ($-1.56$), and \lang{Swa} ($-1.20$). At the rubric level the decrease is largest on Fluency, which is $3.35$--$3.67$ across all resource tiers, while Adequacy differs by at most $0.5$ points (Appendix~\ref{app:rubric_dims}). 

Table~\ref{tab:rq3_gap} reports the English minus low-resource difference per strategy. The overall difference ranges from $-0.130$ for cross-Lingual to $+0.206$ for CoT, the latter with a Math component of $+0.542$. Monolingual is $+0.009$ overall. Excluding SSA-MTE, most MT differences are within $0.11$ of zero.

\begin{table}[htbp]
\centering
\small
\setlength{\tabcolsep}{5pt}
\begin{tabular}{l rrr}
\toprule
\textbf{Task} & \textbf{High} & \textbf{Mid+Low-mid} & \textbf{Low} \\
\midrule
MT  & 0.13 & 0.29 & 1.60 \\
Sum & 0.41 & 0.16 & 0.51 \\
\bottomrule
\end{tabular}
\caption{Cross-lingual range (max$-$min of per-language means) per
resource tier, averaged across strategies. Per-strategy breakdown in
Appendix~\ref{app:heatmaps}.}
\label{tab:rq1_range}
\end{table}

\begin{table}[htbp]
\centering
\small
\setlength{\tabcolsep}{4pt}
\resizebox{\columnwidth}{!}{
    \begin{tabular}{l rrrr r}
    \toprule
    \textbf{Strategy} & \textbf{MT}$^{\ddagger}$ & \textbf{Sum} & \textbf{QA} & \textbf{Math} & \textbf{Overall} \\
    \midrule
    Cross-Ling. & $+$0.091 & $-$0.097 & $-$0.197 & $-$0.789 & $-$0.130 \\
    Monolingual & $+$0.214 & $+$0.092 & $-$0.258 & $-$0.466 & $+$0.009 \\
    Zero-Shot   & $+$0.064 & $-$0.101 & $+$0.245 & $-$0.349 & $+$0.096 \\
    CoT         & --       & $-$0.004 & $+$0.081 & $+$0.542 & $+$0.206 \\
    \bottomrule
    \multicolumn{6}{l}{\footnotesize $^{\ddagger}$ SSA-MTE excluded. -- = excluded by design.}
    \end{tabular}
}
\caption{English--low-resource difference per strategy ($\Delta$ =
English $-$ low-resource, averaged over 4 judges). Positive values
indicate a higher English mean. Full table (9 strategies) in
Appendix~\ref{app:gap}.}
\label{tab:rq3_gap}
\end{table}

\subsection{Agreement with Human Judgments}
\label{sec:validity}

Table~\ref{tab:validity} reports instance-level Spearman correlations against human quality annotations. All judge--strategy pairs correlate at $\rho=0.22$--$0.46$ with $p<0.001$. The cross-dataset mean is highest for Role Prompting ($\rho=0.483$) and lowest for Trans-Eval ($\rho=0.321$). On SSA-MTE, Monolingual is $\rho=0.455$ for Gemma and $\rho=0.219$ for Aya, the highest and lowest values in that column.

SSA-MTE provides the only non-English human annotations, so the Summarization, QA, and Math results are not validated against human judgments in the low-resource languages under study.

\begin{table}[t]
\centering
\small
\setlength{\tabcolsep}{3.5pt}
\begin{tabular}{l cccc r}
\toprule
\textbf{Strategy} & Aya & Gemma & Q32 & Q72 & \textbf{Mean} \\
\midrule
Role        & .404 & .423 & .415 & .366 & \textbf{.483} \\
Cross-Ling. & .326 & .406 & .398 & .335 & .422 \\
Monolingual & .219 & .455 & .414 & .356 & .366 \\
Trans-Eval  & .244 & .331 & .266 & .250 & .321 \\
\bottomrule
\multicolumn{6}{l}{\footnotesize Full table (7 strategies) in Appendix~\ref{app:validity}.}
\end{tabular}
\caption{Instance-level Spearman $\rho$ against human scores on
SSA-MTE, and cross-dataset mean over SSA-MTE, SummEval, and WMT MQM.
All $p<0.001$.}
\label{tab:validity}
\end{table}

\section{Discussion}
\label{sec:discussion}

We find that the effect of a prompting strategy on evaluation scores depends on both the Judge and the target language. Judge dependence is quantified by the Judge $\times$ strategy interactions, exceeding the strategy's effect when measured independently. This, however, comes with an exception on Summarization, where we observe that different judges exhibit different preferences w.r.t.\ the strategies. 

\paragraph{Language effect.} Language dependence is quantified by $\Delta_{\mathrm{mono}}$, which is positive for one judge and negative for another on identical content. Note that we observe an-order-of-magnitude wider range of $\Delta_{\mathrm{mono}}$ for low-resource languages in the cross-lingual setting. 
Rescaling does not affect the nature of the observed dependences, which is in line with \citet{bellibatlu2026judgesense} and extends it from within-judge instability under paraphrase to between-judge non-transfer of complete strategies. The fact that the LLM used as the Judge accounts for the larger share of variance is consistent with multilingual evaluation behaviour being determined by properties acquired in pre-training and instruction tuning \citep{ustun2024aya, gemma3_2025}, rather than by prompt design, though our design does not isolate these sources from other inter-model differences.

\paragraph{Strategy effect.} The choice of Strategy has lower effects than the Judge LLM. Nevertheless, it is still not negligible. We observe the strongest Strategy effect for QA and Summarization tasks ($\eta^2=0.059$ \& $0.057$, respectively)
with open-ended semantics where prompt framing shapes which quality
aspects the judge attends to. Chain-of-Thought shows the lowest
cross-lingual variability (mean CV $=0.036$), consistent with prior
work, suggesting structured reasoning reduces variance by making
intermediate steps explicit~\cite{wei2022chain,wang2023selfconsistency},
though it does not consistently achieve the highest human alignment,
indicating consistency does not guarantee quality. 

\paragraph{Target-language prompting} illustrates an
architecture-dependent effect: it benefits Gemma ($\Delta=+0.83$) but
harms Aya ($\Delta=-0.32$), with severe degradation for low-resource
African languages (e.g.\ \lang{Yor}: $-2.16$, \lang{Hau}: $-1.56$). Since this holds on
identical languages, it likely reflects judge architecture rather than
data characteristics, possibly because Aya, trained explicitly for
multilingual instruction following~\cite{ustun2024aya}, has developed
multilingual processing that target-language prompts disrupt rather
than support, while Gemma relies more on prompt language during
evaluation, supported by Monolingual achieving Gemma's highest human
alignment ($\rho=0.455$) but Aya's lowest ($\rho=0.219$).

\paragraph{Resource level} is associated with increased instability throughout:
cross-lingual performance ranges widen as language resources decrease, and inter-model rank
agreement drops from moderate in high-resource MT ($\rho=0.46$) to
near zero in low-resource MT ($\rho=0.04$), consistent with the findings of \citet{fuliu2025} and \citet{chung2026crosslingual}, extending their
findings by showing instability affects strategy \emph{rankings}, not
only absolute scores. 

\paragraph{Language bias.} Strategy choice also carries fairness
implications, not obvious from mean scores: Cross-lingual is the most
balanced strategy overall ($\Delta=-0.130$), while CoT shows the
largest English bias in Maths ($\Delta=+0.542$), and aggregate reporting
can mask this variability. Note that Monolingual's near-zero overall gap
($\Delta=+0.009$) hides large positive and negative per-language
biases.

\paragraph{Upshot.}
Judge selection appears more important than prompt selection overall;
however, for some tasks the judge $\times$ strategy interaction contributes
as much as the Judge LLM alone, meaning a prompt optimised for one
judge does not necessarily transfer to another, and strategies should
be validated on the target judge rather than adopted from a different
architecture. Reliability should also be assessed at the language
level, since aggregate metrics can hide substantial cross-language
variation. Among evaluated strategies, Role Prompting offers the best
overall trade-off, achieving the highest human alignment
($\rho=0.483$) while remaining relatively stable across languages.
Improving multilingual LLM-as-a-Judge reliability therefore requires
the combined choice of judge model, prompting strategy, and target
languages, not prompt design alone.

\section{Conclusion}
\label{sec:conclusion}

We evaluated nine prompting strategies across four open-weight judges, 27 languages, and four generation tasks. Prompting effects varied across judges: the judge--strategy interaction exceeded the strategy main effect, judges showed limited agreement on strategy rankings, and target-language prompting produced effects of opposing direction across models. The strategy with the lowest cross-language dispersion also differed from that with the highest aggregate correlation with human judgements, indicating that score stability does not imply evaluation validity. Prompt design should therefore be treated as part of the judge configuration and validated for the intended judge, task, and language. Future work should evaluate additional judge families and system-generated outputs spanning a wider quality range, quantify the effects of prompting on system rankings, and develop prompt-selection methods that improve agreement with human judgements across languages.

\section*{Limitations}
\label{sec:limitations}

\paragraph{Evaluation design.} Gold-reference evaluation isolates
judge behaviour from output variation but introduces ceiling effects
in QA and reference-based MT, limiting strategy discrimination for
those tasks (Section~\ref{sec:rq1}); we confirm this directly using
AfroBench real system outputs but only for a subset of judges, tasks,
and strategies. A related consequence is that Fleiss' $\kappa$
collapses toward zero on the skewed gold-reference score distribution
despite substantial raw agreement, an artefact of chance-correction
under skew rather than genuine disagreement (Appendix~\ref{app:ceiling}).

\paragraph{Language coverage and verification.} Target-language
prompts were native-speaker-verified for only a subset of languages;
part of the observed Monolingual effect may therefore reflect prompt
quality rather than judge behaviour alone. Our language distribution
is weighted toward low-resource languages, so cross-tier comparisons
should be interpreted with caution, and our two-way ANOVA does not
capture language interactions, which we analyse separately via
one-way ANOVAs.

\paragraph{Validation and model coverage.} SummEval and WMT MQM are
English-only, so SRQ4 is fully tested for MT (via SSA-MTE) but not for
non-English summarisation or reasoning. We evaluate four open-source
judges; our findings on judge-specificity may not generalise to
closed-source models, which we test only in a limited, exploratory
capacity (Appendix~\ref{app:ceiling}).

\section*{Ethical considerations.} 
Our results should not be read as
grounds for excluding low-resource languages from LLM-as-a-judge
pipelines; rather, they show that judge and strategy choice must be
validated per language before deployment, particularly where
evaluation outcomes affect downstream model selection or content
moderation for these communities. All datasets used are publicly
released for research use, and we release our own scores under the
same principle to support independent verification. Our experiments
consumed substantial GPU compute (Section~\ref{sec:implementation});
we mitigate this by releasing all 1.07M scores so that others can
reuse them without re-running the benchmark.
\label{sec:label}

\bibliography{references}

\appendix

\section{Additional Methodology Details}
\label{app:language-coverage}

Table~\ref{tab:datasets-full} extends Table~\ref{tab:datasets} from
the main text with the complete per-dataset language codes.

\begin{table*}[!t]
\centering
\small
\setlength{\tabcolsep}{4pt}
\renewcommand{\arraystretch}{0.92}
\begin{tabularx}{\textwidth}{llc p{3.1cm} >{\raggedright\arraybackslash}X}
\toprule
\textbf{Task} & \textbf{Dataset} & \textbf{\#Lang.} & \textbf{Coverage} & \textbf{Languages} \\
\midrule
MT & FLORES-200~\citep{nllb2024nature} & 14
& African and higher-resource languages
& EN, FR, DE, ES, CA, FA, HA, SW, YO, AM, IBO, KIN, XHO, ZUL \\

MT & MAFAND-MT~\citep{adelani2022mafand} & 10
& African languages
& HA, SW, YO, AM, IBO, KIN, PCM, XHO, ZUL, TWI \\

MT & SSA-MTE~\citep{ojo2025ssamte} & 11
& African languages
& HA, SW, YO, AM, IBO, KIN, LUG, SOM, SOT, EWE, XHO \\
\midrule

Summ. & XL-Sum~\citep{hasan2021xlsum} & 13
& African and higher-resource languages
& EN, FR, ES, FA, HA, YO, SW, AM, IBO, SOM, ORM, PCM, TIR \\

Summ. & MLSUM~\citep{scialom2020mlsum} & 2
& European languages
& ES, FR \\

Summ. & CaSum~\citep{casum2022} & 1
& Catalan
& CA \\
\midrule

QA & AfriQA~\citep{ogundepo2023afriqa} & 9
& African languages
& HA, SW, YO, ZUL, KIN, IBO, TWI, BEM, FON \\

QA & XQuAD~\citep{artetxe2020xquad} & 3
& Higher-resource and Arabic
& ES, EN, AR \\

QA & PersianQA~\citep{ayoubi2021persianqa} & 1
& Persian
& FA \\

QA & CatalanQA~\citep{armengol2022catalanqa} & 1
& Catalan
& CA \\

QA & Belebele~\citep{bandarkar2024belebele} & 15
& African languages
& 15 benchmark languages used in our QA subset \\
\midrule

Math & AfriMGSM~\citep{adelani2024afrimgsm} & 16
& African languages
& HA, SW, YO, AM, IBO, KIN, LUG, XHO, ZUL, EWE, LIN, ORM, SOT, WOL, TWI, PCM \\

Math & MGSM~\citep{shi2022mgsm} & 4
& Higher-resource languages
& EN, ES, FR, DE \\
\bottomrule
\end{tabularx}
\caption{Datasets, tasks, language coverage, and language codes.
Language codes follow the conventions used in each source dataset.}
\label{tab:datasets-full}
\end{table*}

\section{Task-Specific Rubric Definitions}
\label{app:rubrics}

All rubrics are included in the system prompt for every 
strategy and judge model. Each dimension is scored 
independently on a 1--5 scale, followed by a single 
holistic overall score used for cross-task comparison.

\subsection*{Machine Translation}
\textit{MQM framework~\cite{freitag2021experts}}

\noindent\textbf{Dimension 1: Adequacy} (meaning preservation)
\begin{itemize}
  \item[\textbf{5}] All meaning from the source preserved; no omissions or additions.
  \item[\textbf{4}] Core meaning correct; minor omissions not affecting comprehension.
  \item[\textbf{3}] General idea captured but notable information missing or inaccurate.
  \item[\textbf{2}] Significant meaning distortion; key information missing or mistranslated.
  \item[\textbf{1}] Meaning completely lost; translation unrelated to source.
\end{itemize}

\noindent\textbf{Dimension 2: Fluency} (naturalness in target language)
\begin{itemize}
  \item[\textbf{5}] Native-like fluency; no grammatical errors; natural tone and register.
  \item[\textbf{4}] Mostly natural with minor awkwardness.
  \item[\textbf{3}] Understandable but noticeably non-native phrasing.
  \item[\textbf{2}] Difficult to read; frequent grammatical errors.
  \item[\textbf{1}] Incomprehensible or gibberish.
\end{itemize}

\noindent\textbf{Dimension 3: Terminology}
\begin{itemize}
  \item[\textbf{5}] All domain-specific terms translated correctly and consistently.
  \item[\textbf{4}] Most terms correct; minor inconsistencies.
  \item[\textbf{3}] Some key terms wrong but meaning partially recoverable.
  \item[\textbf{2}] Multiple terminology errors misleading the reader.
  \item[\textbf{1}] Key terminology completely wrong or absent.
\end{itemize}

\subsection*{Summarization}
\textit{SEAHORSE framework~\cite{clark2023seahorse,fabbri2021summeval}}

\noindent\textbf{Dimension 1: Comprehensibility}
\begin{itemize}
  \item[\textbf{5}] Perfectly clear and understandable on its own.
  \item[\textbf{4}] Mostly clear; minor ambiguity.
  \item[\textbf{3}] Understandable with effort; some unclear passages.
  \item[\textbf{2}] Difficult to understand; significant clarity issues.
  \item[\textbf{1}] Incoherent or incomprehensible.
\end{itemize}

\noindent\textbf{Dimension 2: Attribution} (factual grounding in source)
\begin{itemize}
  \item[\textbf{5}] Every claim traceable to the source article.
  \item[\textbf{4}] Almost all claims supported; one minor unsupported detail.
  \item[\textbf{3}] Most claims supported but some fabrications.
  \item[\textbf{2}] Significant fabricated content.
  \item[\textbf{1}] Entirely fabricated or contradicts the source.
\end{itemize}

\noindent\textbf{Dimension 3: Main Ideas} (coverage)
\begin{itemize}
  \item[\textbf{5}] All key points of the source captured.
  \item[\textbf{4}] Most key points covered; one minor omission.
  \item[\textbf{3}] Some key points covered but important information missing.
  \item[\textbf{2}] Only peripheral details; main points missing.
  \item[\textbf{1}] Misses all important content.
\end{itemize}

\noindent\textbf{Dimension 4: Conciseness}
\begin{itemize}
  \item[\textbf{5}] Optimal length; no redundancy, no critical omissions.
  \item[\textbf{4}] Slightly verbose or brief but acceptable.
  \item[\textbf{3}] Noticeably too long or too short.
  \item[\textbf{2}] Very verbose with repetition, or critically short.
  \item[\textbf{1}] Excessively long or short to be unusable.
\end{itemize}

\subsection*{QA with Passage}
\textit{MM-Eval framework~\cite{son2024mmeval,kim2024prometheus}.
Applied to XQuAD, PersianQA, CatalanQA, Belebele.}

\noindent\textbf{Dimension 1: Relevance}
\begin{itemize}
  \item[\textbf{5}] Answer directly and precisely addresses the question asked.
  \item[\textbf{4}] Mostly relevant; minor tangential content.
  \item[\textbf{3}] Partially relevant; addresses some but not the core of the question.
  \item[\textbf{2}] Marginally relevant; mostly off-topic.
  \item[\textbf{1}] Completely unrelated to the question.
\end{itemize}

\noindent\textbf{Dimension 2: Completeness}
\begin{itemize}
  \item[\textbf{5}] Fully addresses all aspects of the question.
  \item[\textbf{4}] Addresses the main point; minor aspects not covered.
  \item[\textbf{3}] Partially addresses the question; key aspects missing.
  \item[\textbf{2}] Only tangentially addresses the question.
  \item[\textbf{1}] Does not address the question at all.
\end{itemize}

\noindent\textbf{Dimension 3: Passage Grounding}
\begin{itemize}
  \item[\textbf{5}] Answer is directly extractable from or fully supported by the passage.
  \item[\textbf{4}] Answer is mostly grounded; uses minor inference beyond the passage.
  \item[\textbf{3}] Answer is partially grounded; mixes passage content with external knowledge.
  \item[\textbf{2}] Answer is largely not supported by the passage.
  \item[\textbf{1}] Answer contradicts the passage or ignores it entirely.
\end{itemize}

\subsection*{QA without Passage}
\textit{MM-Eval framework~\cite{son2024mmeval}.
Applied to AfriQA (open-retrieval; no passage provided).
Passage Grounding is excluded by design.}

Uses Relevance and Completeness only, with identical 
anchors to the QA with Passage rubric above.

\subsection*{Mathematical Reasoning}
\textit{Process-reward framework~\cite{lightman2023verify,cobbe2021gsm8k}.
Evaluated as text generation: reasoning quality, not only 
the final answer.}

\noindent\textbf{Dimension 1: Answer Correctness}
\begin{itemize}
  \item[\textbf{5}] Final numerical answer is exactly correct.
  \item[\textbf{4}] Very close; minor rounding or unit issue.
  \item[\textbf{3}] Wrong but the approach was reasonable.
  \item[\textbf{2}] Significantly wrong.
  \item[\textbf{1}] No answer or completely wrong.
\end{itemize}

\noindent\textbf{Dimension 2: Reasoning Validity}
\begin{itemize}
  \item[\textbf{5}] Every step logically sound and mathematically correct.
  \item[\textbf{4}] Mostly sound; one minor logical gap.
  \item[\textbf{3}] Notable gaps but shows some understanding.
  \item[\textbf{2}] Largely flawed or incoherent reasoning.
  \item[\textbf{1}] No reasoning or completely nonsensical.
\end{itemize}

\noindent\textbf{Dimension 3: Solution Completeness}
\begin{itemize}
  \item[\textbf{5}] All necessary steps shown clearly.
  \item[\textbf{4}] Most steps shown; one minor step implicit.
  \item[\textbf{3}] Some steps shown but important work missing.
  \item[\textbf{2}] Only the final answer with minimal work.
  \item[\textbf{1}] Just a number with no work shown.
\end{itemize}
\section{Prompt Examples}
\label{app:prompts}

All strategies include the task-specific rubric in the system prompt.
The complete templates are provided for Zero-Shot
(\S\ref{app:prompt-zero-shot}), Few-Shot Anchored
(\S\ref{app:prompt-few-shot}), Prometheus
(\S\ref{app:prompt-prometheus}), Chain-of-Thought
(\S\ref{app:prompt-cot}), Cross-Lingual
(\S\ref{app:prompt-cross-lingual}), Monolingual Prompting
(\S\ref{app:prompt-monolingual}), Translate-then-Evaluate
(\S\ref{app:prompt-translate-evaluate}), Role Prompting
(\S\ref{app:prompt-role}), and Self-Consistency
(\S\ref{app:prompt-self-consistency}). Unless otherwise stated,
the examples use an English--Swahili machine-translation instance
from FLORES-200. The Monolingual Prompting example uses Catalan.
Self-Consistency is described separately because it applies only to
question answering and mathematical reasoning.
Rubric text is abbreviated here; full definitions are in
Appendix~\ref{app:rubrics}.

\subsection*{Zero-shot}
\label{app:prompt-zero-shot}

\begin{promptbox}[zero-shot]{Zero-Shot Evaluation Prompt}
\small
\textbf{[System]}\\
You are an evaluator of machine translation quality.

\medskip
\textbf{Rubric}\\
Adequacy, Fluency, Terminology --- full criteria.

\medskip
\textbf{[User]}\\
Evaluate the quality of the following machine translation.

\medskip
\textbf{Source text (English)}\\
\texttt{``We now have 4-month-old mice...''}

\medskip
\textbf{Reference translation (Swahili)}\\
\texttt{``Sasa tuna panya...''}

\medskip
\textbf{Model translation}\\
\texttt{\{TEXT TO EVALUATE\}}

\medskip
Score each rubric dimension, then provide an overall score
from 1 to 5 followed by a brief justification.
\end{promptbox}

\subsection*{Few-Shot Anchored}
\label{app:prompt-few-shot}

\begin{promptbox}[few-shot]{Few-Shot Anchored Evaluation Prompt}
\small
\textbf{[System]}\\
You are an evaluator of machine translation quality.

\medskip
\textbf{Rubric}\\
Adequacy, Fluency, Terminology --- full criteria.

\medskip
\textbf{Calibration examples}

\smallskip
\textbf{Example 1 (Score: 5)}\\
\texttt{\{PERFECT TRANSLATION EXAMPLE\}}\\
Adequacy: 5 $\mid$ Fluency: 5 $\mid$ Terminology: 5
$\mid$ Overall: 5

\smallskip
\textbf{Example 2 (Score: 1)}\\
\texttt{\{COMPLETELY INCORRECT TRANSLATION EXAMPLE\}}\\
Adequacy: 1 $\mid$ Fluency: 3 $\mid$ Terminology: 1
$\mid$ Overall: 1

\medskip
\textbf{[User]}\\
Evaluate the quality of the following machine translation.

\medskip
\textbf{Source text (English)}\\
\texttt{\{SOURCE TEXT\}}

\medskip
\textbf{Reference translation}\\
\texttt{\{REFERENCE TRANSLATION\}}

\medskip
\textbf{Model translation}\\
\texttt{\{TEXT TO EVALUATE\}}

\medskip
Score each rubric dimension, then provide an overall score
from 1 to 5 followed by a brief justification.
\end{promptbox}

\subsection*{Prometheus}
\label{app:prompt-prometheus}

\begin{promptbox}[prometheus]{Prometheus Evaluation Prompt}
\small
\textbf{[System]}\\
You are an evaluator of machine translation quality.

\medskip
\textbf{Rubric}\\
Adequacy, Fluency, Terminology --- full criteria.

\medskip
\textbf{Scoring protocol}
\begin{enumerate}[nosep, leftmargin=1.5em]
    \item Score each dimension independently from 1 to 5.
    \item Provide a holistic overall score from 1 to 5.
    \item Report each dimension with its score and a one-sentence
          justification, followed by the overall score.
\end{enumerate}

\medskip
\textbf{[User]}\\
Evaluate the quality of the following machine translation.

\medskip
\textbf{Source text (English)}\\
\texttt{\{SOURCE TEXT\}}

\medskip
\textbf{Reference translation}\\
\texttt{\{REFERENCE TRANSLATION\}}

\medskip
\textbf{Model translation}\\
\texttt{\{TEXT TO EVALUATE\}}
\end{promptbox}

\subsection*{Chain-of-Thought}
\label{app:prompt-cot}

\begin{promptbox}[chain-of-thought]{Chain-of-Thought Evaluation Prompt}
\small
\textbf{[System]}\\
You are an evaluator of machine translation quality.

\medskip
\textbf{Rubric}\\
Adequacy, Fluency, Terminology --- full criteria.

\medskip
Before assigning a rating, reason through the evaluation
step by step:
\begin{enumerate}[nosep, leftmargin=1.5em]
    \item Identify what the task requires.
    \item Score each rubric dimension and explain the score.
    \item Identify the translation's strengths and weaknesses.
    \item Assign the final overall score.
\end{enumerate}

\medskip
\textbf{[User]}\\
Evaluate the quality of the following machine translation.

\medskip
\textbf{Source text (English)}\\
\texttt{\{SOURCE TEXT\}}

\medskip
\textbf{Reference translation}\\
\texttt{\{REFERENCE TRANSLATION\}}

\medskip
\textbf{Model translation}\\
\texttt{\{TEXT TO EVALUATE\}}

\medskip
Reason step by step, then provide your overall score.
\end{promptbox}

\subsection*{Cross-Lingual}
\label{app:prompt-cross-lingual}

\begin{promptbox}[cross-lingual]{Cross-Lingual Evaluation Prompt}
\small
\textbf{[System]}\\
You are an evaluator of machine translation quality.

\medskip
\textbf{Rubric}\\
Adequacy, Fluency, Terminology --- full criteria.

\medskip
The content below is in Swahili. Evaluate it using the same
criteria applied to English content. Do not penalize the output
for being written in a non-English language.

\medskip
\textbf{[User]}\\
Evaluate the quality of the following machine translation.

\medskip
\textbf{Source text (English)}\\
\texttt{\{SOURCE TEXT\}}

\medskip
\textbf{Reference translation (Swahili)}\\
\texttt{\{REFERENCE TRANSLATION\}}

\medskip
\textbf{Model translation (Swahili)}\\
\texttt{\{TEXT TO EVALUATE\}}

\medskip
Score each rubric dimension, then provide an overall score
from 1 to 5 followed by a brief justification.
\end{promptbox}

\subsection*{Monolingual}
\label{app:prompt-monolingual}

\begin{promptbox}[monolingual]{Monolingual Evaluation Prompt: Catalan}
\small
\textbf{[System]}\\
Ets un expert en avaluació de la qualitat de traduccions.

\medskip
\textbf{Rúbrica}\\
Adequació, fluïdesa i terminologia --- criteris complets.

\medskip
\textbf{[User]}\\
Avalua la qualitat de la traducció següent.

\medskip
\textbf{Text original (anglès)}\\
\texttt{\{TEXT ORIGINAL\}}

\medskip
\textbf{Traducció de referència (català)}\\
\texttt{\{TRADUCCIÓ DE REFERÈNCIA\}}

\medskip
\textbf{Traducció del model (català)}\\
\texttt{\{TEXT PER AVALUAR\}}

\medskip
$1=\text{Molt dolent} \mid 5=\text{Excel·lent}$

\medskip
Dona una puntuació de l'1 al 5 i una justificació breu.

\medskip
\textit{Equivalent verified templates were used for French,
Spanish, Catalan, and Persian.}
\end{promptbox}

\subsection*{Translate-then-Evaluate}
\label{app:prompt-translate-evaluate}

\begin{promptbox}[translate-evaluate]{Translate-then-Evaluate Prompt}
\small
\textbf{[System]}\\
You are an evaluator of machine translation quality.

\medskip
\textbf{Rubric}\\
Adequacy, Fluency, Terminology --- full criteria.

\medskip
Complete the evaluation in two steps:
\begin{enumerate}[nosep, leftmargin=1.5em]
    \item Mentally translate the Swahili output into English and
          state your English understanding.
    \item Evaluate the output based on this English understanding,
          applying all rubric criteria.
\end{enumerate}
Show both steps before reporting the final score.

\medskip
\textbf{[User]}\\
Evaluate the quality of the following machine translation.

\medskip
\textbf{Source text (English)}\\
\texttt{\{SOURCE TEXT\}}

\medskip
\textbf{Reference translation (Swahili)}\\
\texttt{\{REFERENCE TRANSLATION\}}

\medskip
\textbf{Model translation (Swahili)}\\
\texttt{\{TEXT TO EVALUATE\}}
\end{promptbox}

\subsection*{Role Prompting}
\label{app:prompt-role}

\begin{promptbox}[role-prompting]{Role-Based Evaluation Prompt}
\small
\textbf{[System]}\\
You are a professional translator and linguistic evaluator
with extensive experience in multilingual translation-quality
assessment. You are trained in MQM and specialize in
English--Swahili evaluation.

\medskip
\textbf{Rubric}\\
Adequacy, Fluency, Terminology --- full criteria.

\medskip
\textbf{[User]}\\
Evaluate the quality of the following machine translation.

\medskip
\textbf{Source text (English)}\\
\texttt{\{SOURCE TEXT\}}

\medskip
\textbf{Reference translation (Swahili)}\\
\texttt{\{REFERENCE TRANSLATION\}}

\medskip
\textbf{Model translation (Swahili)}\\
\texttt{\{TEXT TO EVALUATE\}}

\medskip
Score each rubric dimension, then provide an overall score
from 1 to 5 followed by a brief justification.
\end{promptbox}

\subsection*{Self-Consistency}
\label{app:prompt-self-consistency}

\begin{promptbox}[self-consistency]{Self-Consistency Evaluation Prompt}
\small
The Chain-of-Thought evaluation prompt is executed independently
three times with a sampling temperature of $0.7$.

\medskip
For each execution, the model:
\begin{enumerate}[nosep, leftmargin=1.5em]
    \item identifies the task requirements;
    \item scores and explains each rubric dimension;
    \item identifies strengths and weaknesses; and
    \item assigns a final overall score.
\end{enumerate}

\medskip
The final score is determined by majority vote. When all three
scores differ, the arithmetic mean is used.

\medskip
\textit{This strategy is applied only to question answering and
mathematical reasoning tasks.}
\end{promptbox}

\section{Cross-Language CV by Strategy}
\label{app:cv_full}
\begin{table*}[h]
    \centering
\setlength{\tabcolsep}{4.5pt}
\begin{tabular}{l rrrr r r rr rr}
\toprule
& \multicolumn{5}{c}{\textbf{CV by judge model}}
& & \multicolumn{2}{c}{\textbf{Resource tier}}
& \multicolumn{2}{c}{\textbf{Task}} \\
\cmidrule(lr){2-6} \cmidrule(lr){8-9} \cmidrule(lr){10-11}
\textbf{Strategy}
& Aya & Gemma & Q32 & Q72 & Mean
& & High & Low & MT & Sum \\
\midrule
\textbf{CoT}
  & 0.070 & \textbf{0.024} & \textbf{0.012} & \textbf{0.039} & \textbf{0.036}
  && \textbf{0.030} & \textbf{0.009} & ---            & \textbf{0.021} \\
Trans-Eval
  & 0.078 & 0.105 & 0.137 & 0.092 & 0.103
  && 0.080          & 0.081          & 0.086          & 0.041 \\
Cross-Ling.
  & 0.080 & 0.118 & 0.140 & 0.151 & 0.122
  && 0.057          & 0.120          & 0.124          & 0.030 \\
Role
  & 0.093 & 0.123 & 0.140 & 0.160 & 0.129
  && 0.062          & 0.125          & 0.129          & 0.032 \\
Few-Shot
  & 0.092 & 0.116 & 0.141 & 0.181 & 0.133
  && 0.059          & 0.138          & 0.140          & 0.025 \\
Zero-Shot
  & 0.087 & 0.134 & 0.146 & 0.173 & 0.135
  && 0.075          & 0.133          & 0.137          & 0.031 \\
\textit{Monolingual}
  & \textit{0.186} & 0.114 & 0.163 & 0.184 & \textit{0.162}
  && \textit{0.166} & 0.122          & 0.132          & \textit{0.099} \\
\textit{Prometheus}
  & 0.107 & \textit{0.236} & 0.140 & 0.176 & \textit{0.165}
  && 0.059          & 0.115          & \textbf{0.117} & 0.033 \\
\bottomrule
\end{tabular}
\caption{\textbf{Cross-language CV per strategy (MT + Summarization),
by judge model, resource tier, and task.}
CV $= \sigma/\mu$ of per-language means; lower = more consistent.
High and Low resource tiers only; Mid and Low-mid excluded ($n=2$
languages each, insufficient for reliable CV estimation).
\textbf{Bold} = most stable per column; \textit{italic} = least stable.}
\label{tab:cv_full}
\end{table*}

\section{Per-Model Strategy $\times$ Dataset Heatmaps}
\label{app:heatmaps}

\begin{figure*}[!t]
  \centering
  \includegraphics[width=\textwidth]{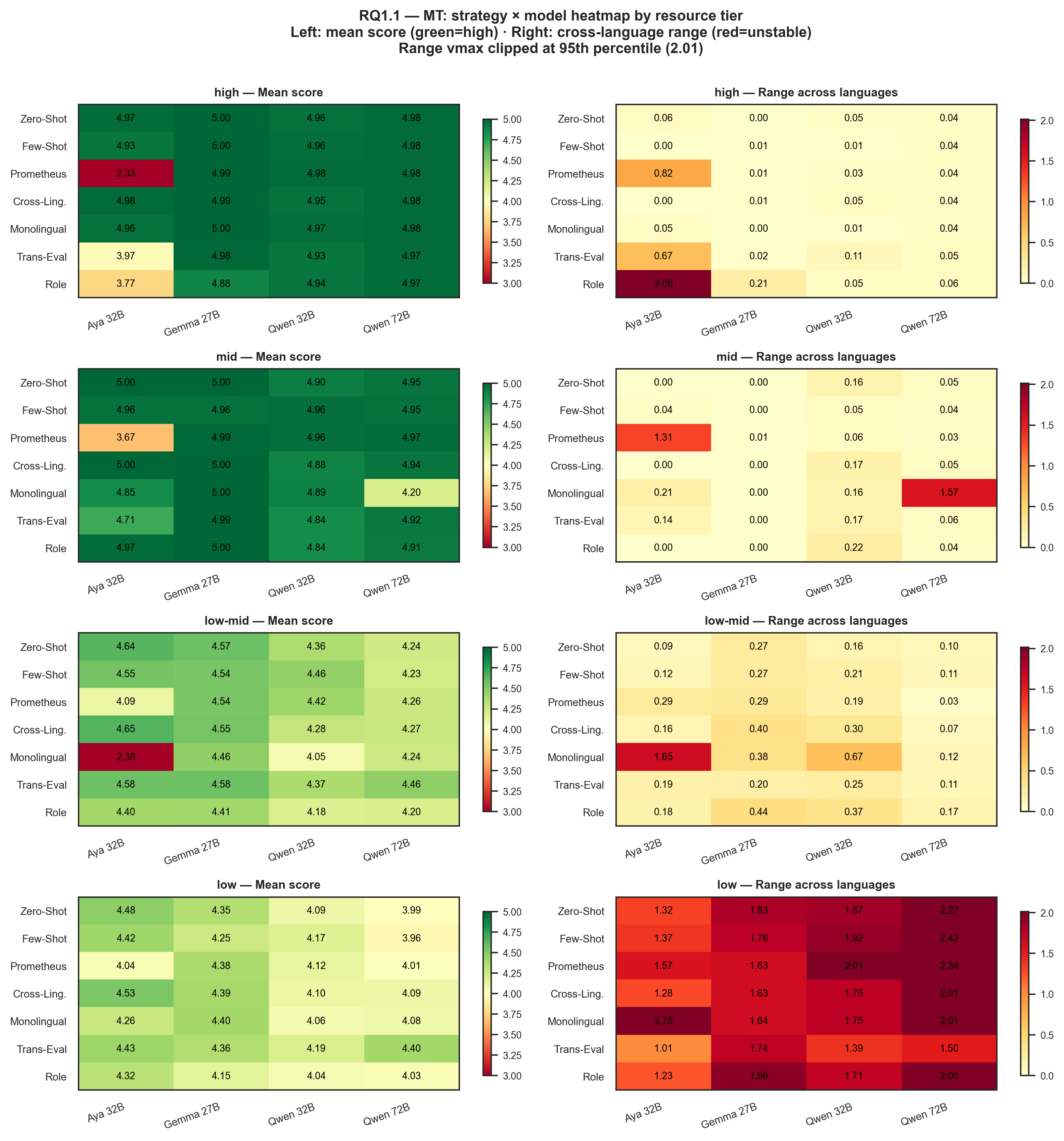}
  \caption{\textbf{MT: mean score (left) and cross-language
  range (right) per strategy, judge model, and resource tier.}
  High-resource panels show near-uniform green (ceiling effect
  on reference-based MT); the range column reveals large
  inconsistencies for low-resource languages (1.41--2.04)
  regardless of strategy, consistent with our finding
  that MT inconsistency is a resource-level phenomenon.}
  \label{fig:heatmap_mt}
\end{figure*}

\begin{figure*}[!t]
  \centering
  \includegraphics[width=\textwidth]{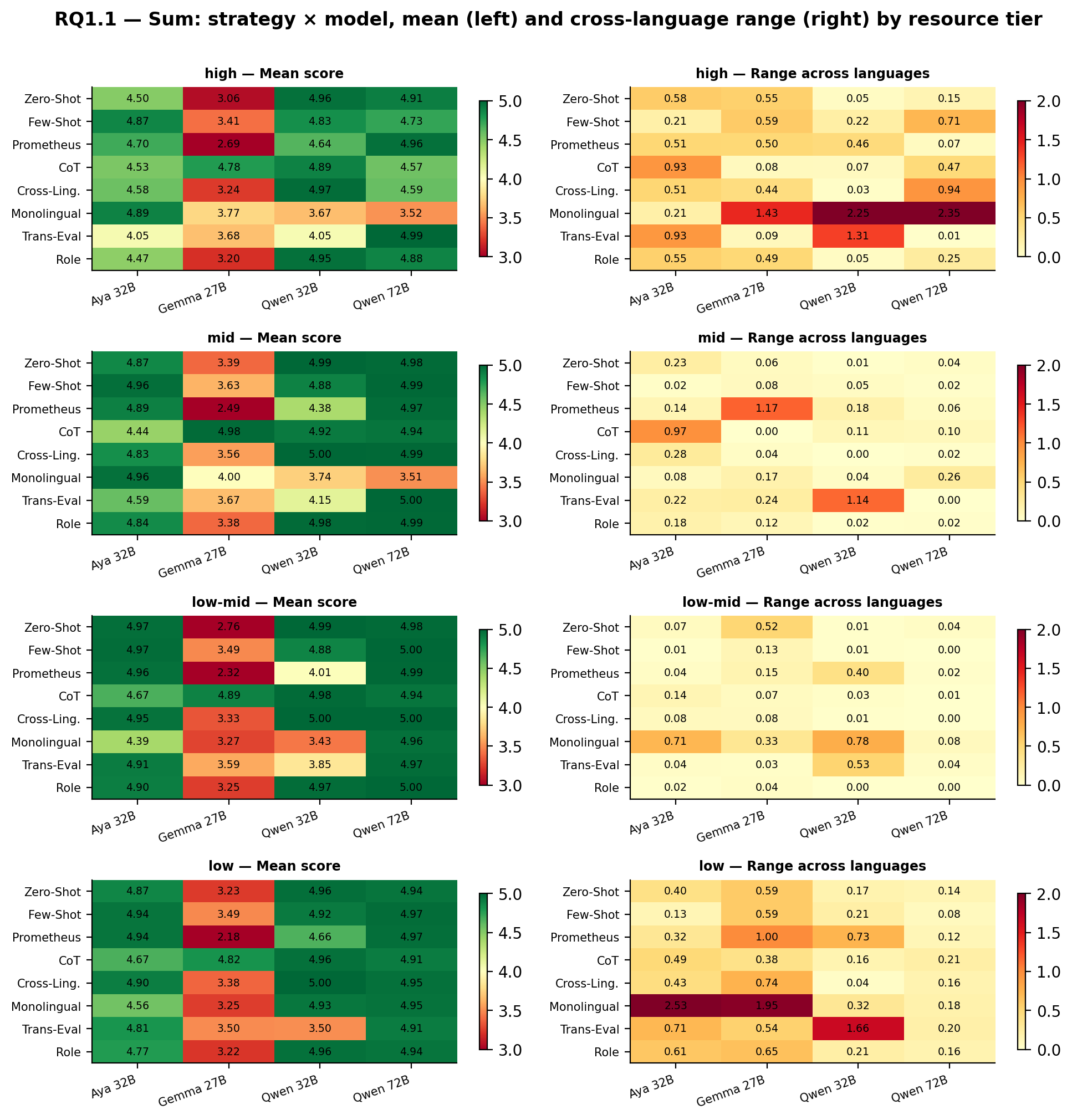}
  \caption{\textbf{Summarization: mean score (left) and
  cross-language range (right) per strategy, judge model,
  and resource tier.}
  Monolingual shows the highest range for Aya 32B across
  all resource tiers, while CoT and Role show consistently
  low range. Strategy effects on Summarization scoring are
  substantially model-dependent.}
  \label{fig:heatmap_sum}
\end{figure*}

\begin{figure*}[!t]
  \centering
  \includegraphics[width=\textwidth]{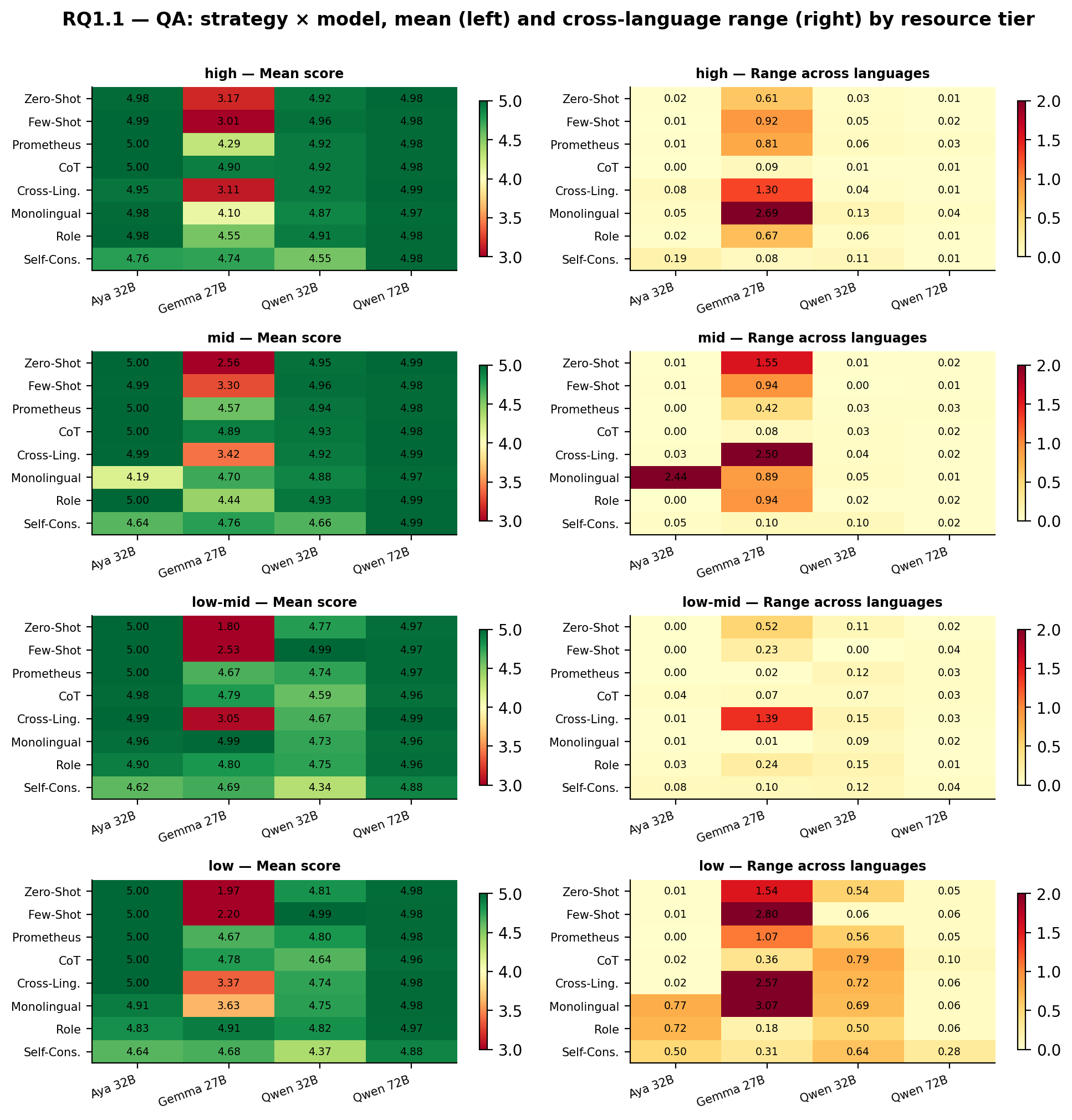}
  \caption{\textbf{QA: mean score (left) and cross-language
  range (right) per strategy, judge model, and resource tier.}
  Near-uniform green across all panels reflects the ceiling
  effect from gold-reference evaluation; range values are
  low throughout, consistent with the ceiling interpretation.}
  \label{fig:heatmap_qa}
\end{figure*}

\begin{figure*}[!t]
  \centering
  \includegraphics[width=\textwidth]{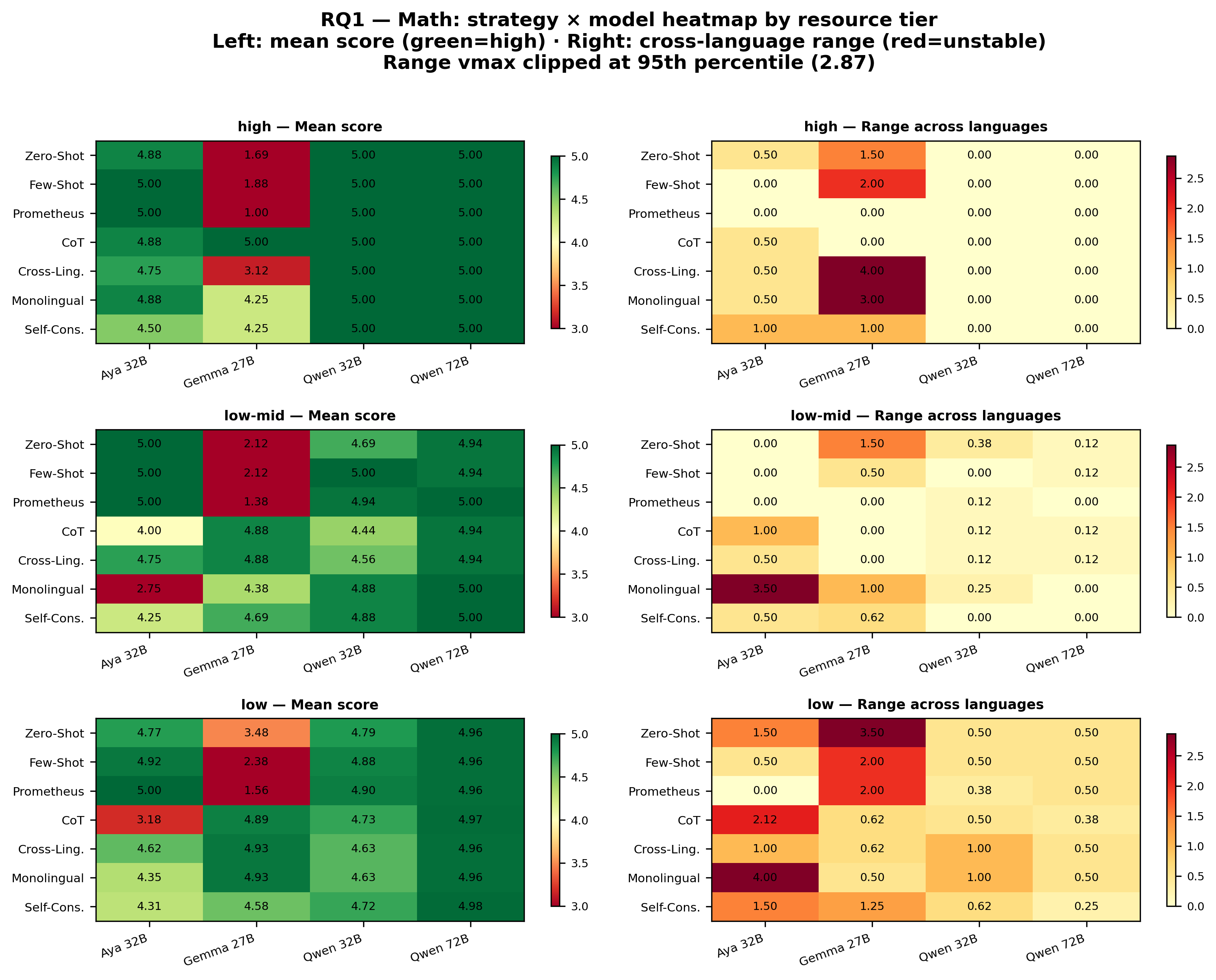}
  \caption{\textbf{Math: mean score (left) and cross-language
  range (right) per strategy, judge model, and resource tier.}
  Gemma 27B shows substantially lower Zero-Shot scores
  (CoT rescue effect visible in mean panels) and higher
  range under Monolingual for low-resource languages.
  Math mid-resource is empty: AfriMGSM covers only low
  and low-mid languages.}
  \label{fig:heatmap_math}
\end{figure*}

Figures~\ref{fig:heatmap_mt}, \ref{fig:heatmap_sum}, \ref{fig:heatmap_qa}
and \ref{fig:heatmap_math} provide the full per-model breakdown of mean
score and cross-language range per strategy and resource tier across
all four tasks, extending the MT and Summarization summary shown in
Figure~\ref{fig:rq1_overview} in the main text.

\section{Full Inter-Model Rank Agreement }
\label{app:intermodel_full}

\begin{figure*}[!t]
  \centering
  \includegraphics[width=0.8\textwidth]{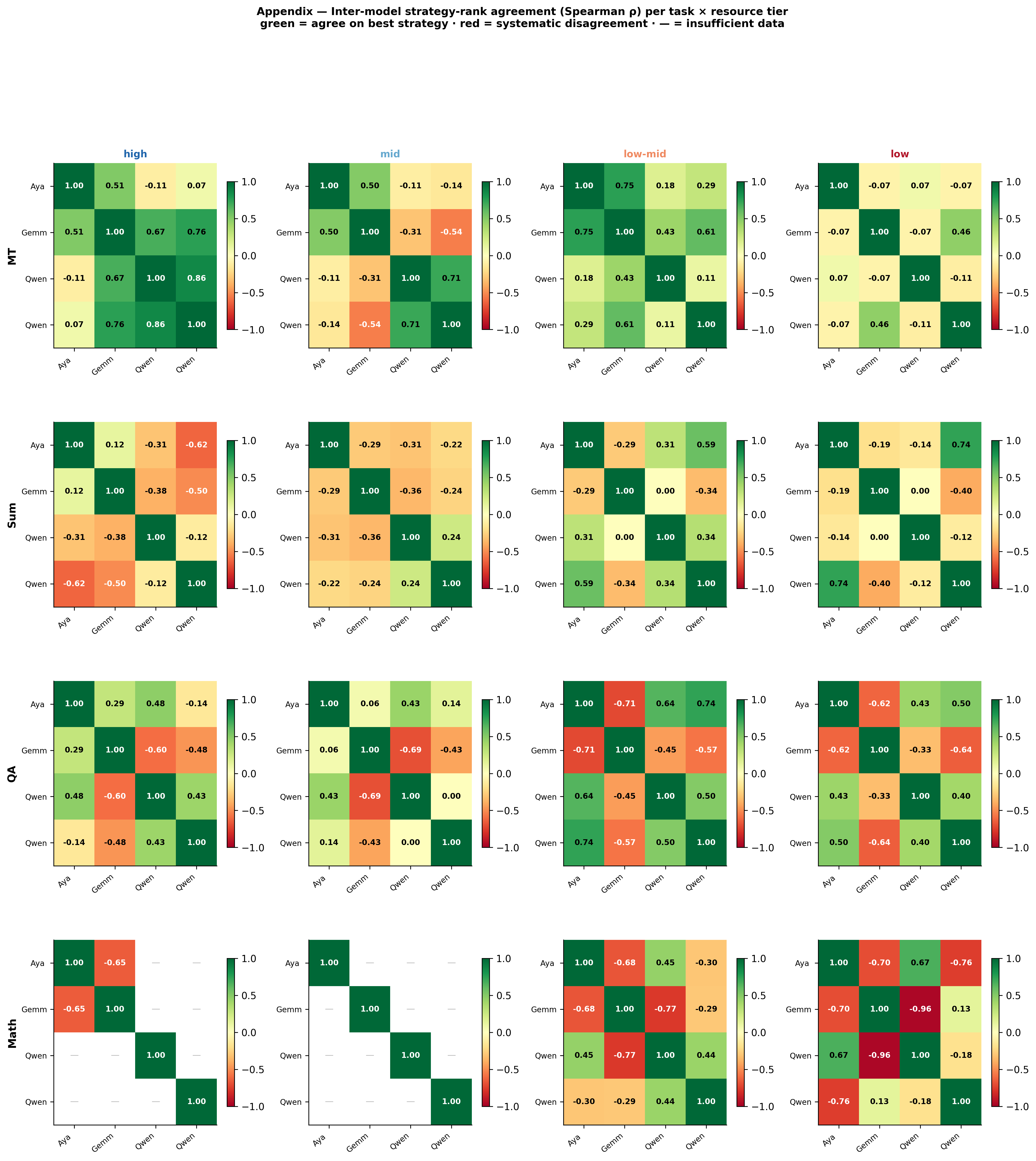}
  \caption{\textbf{Inter-model strategy-rank agreement
  (Spearman $\rho$) per model pair, task, and resource tier.}
  Each $4\times4$ matrix shows pairwise $\rho$ for one
  task $\times$ resource combination.
  Green = models agree on best strategy; red = systematic
  disagreement; --- = insufficient data.
  The averaged values per task and resource tier are
  reported in the main text.
  Gemma 27B systematically disagrees with all other models
  for Summarization, consistent with its divergent
  zero-shot scoring scale.}
  \label{fig:appendix_spearman}
\end{figure*}

\section{Rubric Dimension Analysis}
\label{app:rubric_dims}

\begin{figure*}[!t]
  \centering
  \includegraphics[width=\textwidth]{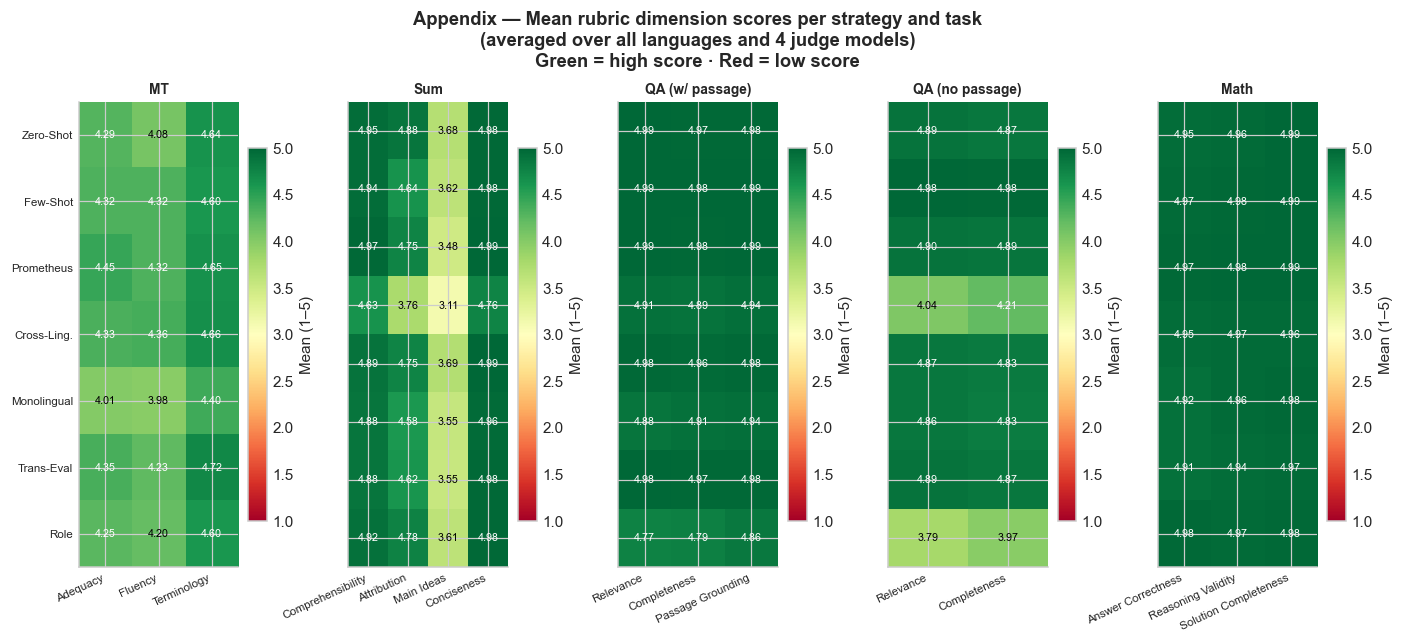}
  \caption{\textbf{Mean rubric dimension scores per strategy
  and task (averaged over all languages and 4 judge models).}
  Green = high score; red = low score (scale 1--5).
  MT Terminology and Summarization Main Ideas are the most
  strategy-sensitive dimensions; QA and Math dimensions
  show near-ceiling values consistent with the gold-reference
  evaluation design. QA is split into passage-based
  (XQuAD, PersianQA, CatalanQA, Belebele) and passage-free
  (AfriQA) variants due to differing rubric dimensions.}
  \label{fig:rubric_overall}
\end{figure*}

\begin{figure*}[!t]
  \centering
  \includegraphics[width=0.9\textwidth]{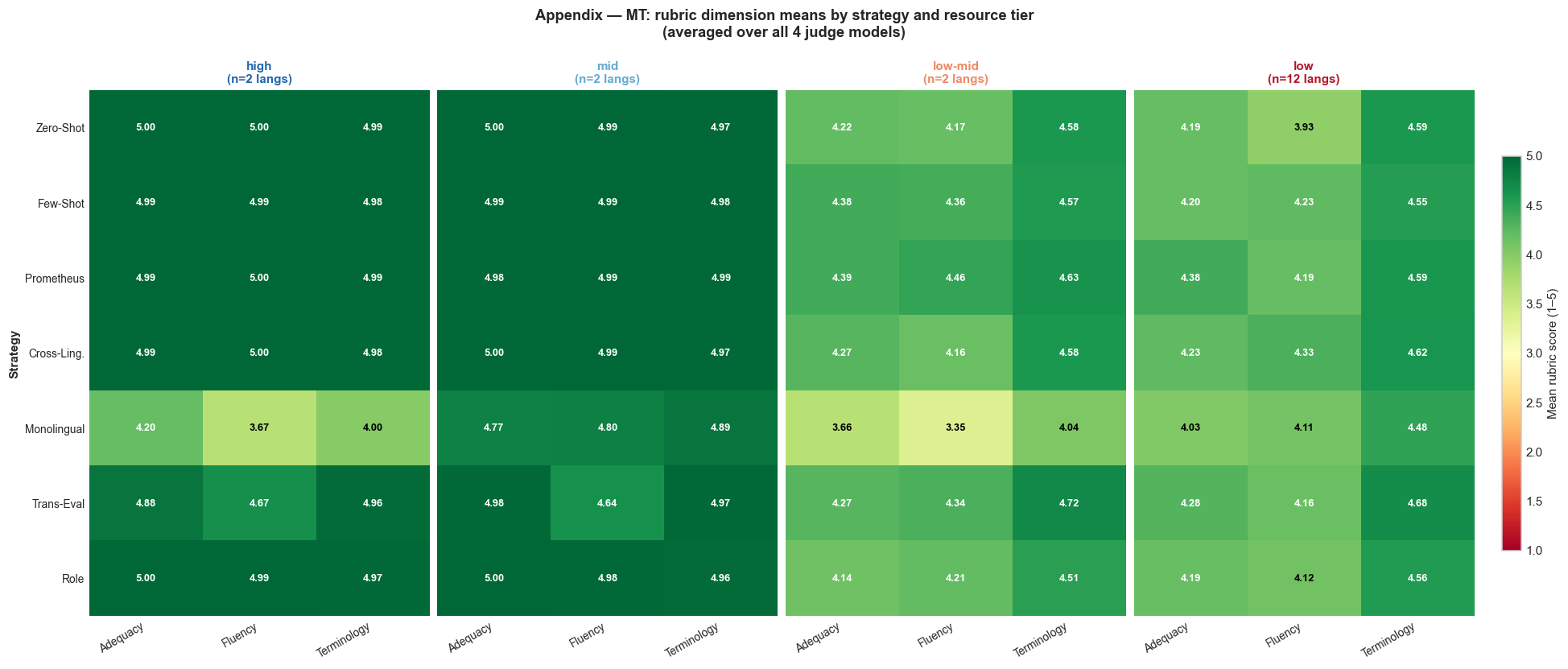}
  \caption{\textbf{MT rubric dimension means by strategy and
  resource tier (averaged over 4 judge models).}
  The resource-level effect concentrates in Terminology
  (3.93--4.59 for low-resource vs 4.97--5.00 for
  high-resource) while Adequacy and Fluency remain stable,
  confirming that judges detect terminological inconsistency
  as the primary quality signal for low-resource MT.}
  \label{fig:rubric_mt_resource}
\end{figure*}

\begin{figure*}[!t]
  \centering
  \includegraphics[width=0.9\textwidth]{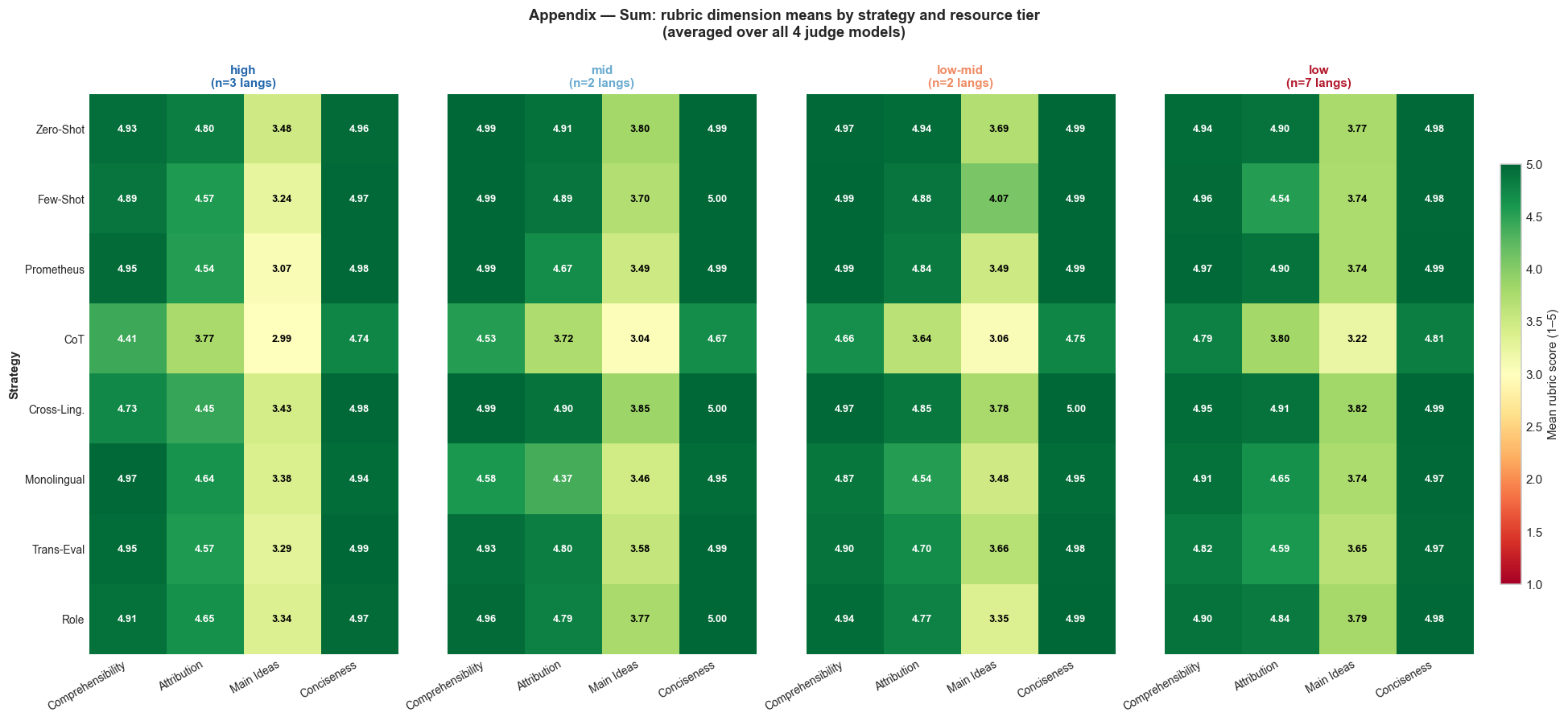}
  \caption{\textbf{Summarization rubric dimension means by
  strategy and resource tier (averaged over 4 judge models).}
  Main Ideas drops to 2.99--3.07 under CoT and Prometheus
  vs 3.48--3.80 for Zero-Shot, consistently across all
  resource tiers, confirming the pattern is a
  rubric--strategy interaction rather than a cross-lingual
  phenomenon.}
  \label{fig:rubric_resource}
\end{figure*}

Figures~\ref{fig:rubric_overall}--\ref{fig:rubric_resource} 
provide the rubric-level evidence discussed in 
Section~\ref{sec:discussion}. MT Terminology and 
Summarization Main Ideas are the most strategy- and 
resource-sensitive dimensions respectively.

\section{Full Variance Decomposition Table}
\label{app:eta}

Table~\ref{tab:eta_squared_full} extends Table~\ref{tab:eta_squared}
from the main text with the MT-excluding-SSA-MTE column omitted
there for space, isolating the reference-based MT effect.

\begin{table*}[!t]
\centering
\small
\setlength{\tabcolsep}{5pt}
\begin{tabular}{l r rrrr r}
\toprule
\textbf{Factor} & \textbf{Overall} & \textbf{MT} &
\textbf{MT}$^{\ddagger}$ &
\textbf{Sum} & \textbf{QA} & \textbf{Math} \\
\midrule
Judge model          & 0.087 & 0.006 & 0.048 & 0.438 & 0.251 & 0.222 \\
Strategy             & 0.012 & 0.007 & 0.035 & 0.057 & 0.059 & 0.036 \\
Judge $\times$ Strat & 0.056 & 0.017 & 0.114 & 0.165 & 0.247 & 0.297 \\
Language$^\dagger$   & 0.043 & 0.325 & 0.030 & 0.032 & 0.012 & 0.010 \\
Resource$^\dagger$   & 0.007 & 0.051 & ---   & 0.010 & 0.003 & 0.001 \\
\bottomrule
\multicolumn{7}{l}{$^\dagger$ One-way ANOVA.\quad
$^\ddagger$ SSA-MTE excluded (reference-based MT only).}
\end{tabular}
\caption{Full variance decomposition ($\eta^2$) of judge scores,
including the MT column with SSA-MTE excluded. All effects
significant ($p<0.001$).}
\label{tab:eta_squared_full}
\end{table*}

\section{Full English--Low-Resource Gap Table}
\label{app:gap}

Table~\ref{tab:rq3_gap_full} extends Table~\ref{tab:rq3_gap} from
the main text with all nine strategies and the Math column.

\begin{table*}[!t]
\centering
\small
\setlength{\tabcolsep}{4pt}
\begin{tabular}{l rrrr r r}
\toprule
\textbf{Strategy} & \textbf{MT} & \textbf{MT}$^{\ddagger}$
& \textbf{Sum} & \textbf{QA} & \textbf{Math}
& \textbf{Overall} \\
\midrule
Cross-Ling. & $+$0.564 & $+$0.091 & $-$0.097 & $-$0.197 & $-$0.789 & $-$0.130 \\
Monolingual & $+$0.670 & $+$0.214 & $+$0.092 & $-$0.258 & $-$0.466 & $+$0.009 \\
Role        & $+$0.088 & $-$0.378 & $-$0.015 & $-$0.116 & NaN       & $-$0.014 \\
Prometheus  & $-$0.069 & $-$0.535 & $+$0.225 & $+$0.030 & $-$0.115 & $+$0.018 \\
Trans-Eval  & $+$0.211 & $-$0.113 & $-$0.004 & NaN       & NaN       & $+$0.104 \\
Zero-Shot   & $+$0.587 & $+$0.064 & $-$0.101 & $+$0.245 & $-$0.349 & $+$0.096 \\
Few-Shot    & $+$0.620 & $+$0.105 & $+$0.009 & $+$0.056 & $-$0.276 & $+$0.102 \\
Self-Cons.  & NaN       & NaN       & NaN       & $+$0.083 & $+$0.096 & $+$0.090 \\
CoT         & NaN       & NaN       & $-$0.004 & $+$0.081 & $+$0.542 & $+$0.206 \\
\bottomrule
\multicolumn{7}{l}{$^{\ddagger}$ SSA-MTE excluded
(FLORES-200 and MAFAND-MT only).}
\end{tabular}
\caption{Full English--low-resource scoring gap per strategy and
task ($\Delta$ = English mean $-$ low-resource mean, averaged
across 4 judge models). Positive values favour English.
MT$^{\ddagger}$ excludes SSA-MTE. Single-task languages excluded
throughout.}
\label{tab:rq3_gap_full}
\end{table*}

\section{Ceiling Effect Confirmation: Gold vs.\ AfroBench}
\label{app:ceiling}

To confirm that compressed scores on gold-reference tasks reflect a
genuine ceiling effect rather than low judge sensitivity, we evaluate
a subset of judges on AfroBench real system outputs
\cite{ojo-etal-2025-afrobench} for MT and Summarization, using a matched set
of five strategies (Zero-Shot, CoT, Role Prompting, Monolingual,
Monolingual-CoT). Table~\ref{tab:ceiling} compares mean and standard
deviation between gold references and AfroBench outputs.

\begin{table*}[!t]
\centering
\small
\setlength{\tabcolsep}{5pt}
\begin{tabular}{l rr rr}
\toprule
& \multicolumn{2}{c}{\textbf{Gold}} & \multicolumn{2}{c}{\textbf{AfroBench}} \\
\cmidrule(lr){2-3} \cmidrule(lr){4-5}
\textbf{Judge} & Mean & SD & Mean & SD \\
\midrule
Aya 32B     & 4.35 & 0.68 & 3.68 & 1.41 \\
Qwen 32B    & 4.42 & 0.61 & 3.35 & 1.03 \\
Qwen 72B    & 4.31 & 0.72 & 3.32 & 1.18 \\
Gemma 27B   & 4.02 & 0.89 & ---  & ---  \\
Gemini 2.5$^\dagger$ & --- & --- & 3.20 & 1.87 \\
\bottomrule
\multicolumn{5}{l}{$^\dagger$ Gemini evaluated on AfroBench only (MT); included as}\\
\multicolumn{5}{l}{robustness check for the ceiling effect.}
\end{tabular}
\caption{Mean judge score (gold vs.\ AfroBench), pooled across MT
and Summarization. AfroBench scores are consistently lower and more
dispersed, confirming the ceiling effect.}
\label{tab:ceiling}
\end{table*}

Every judge shows a lower mean and higher standard deviation on
AfroBench than on gold, confirming that the compression observed on
gold-reference tasks (Section~\ref{sec:rq1}) reflects the evaluation
design rather than an inherent inability of judges to discriminate
quality.

\section{External Validity --- SummEval and WMT MQM}
\label{app:validity}

\begin{figure*}[!t]
  \centering
  \includegraphics[width=0.9\textwidth]{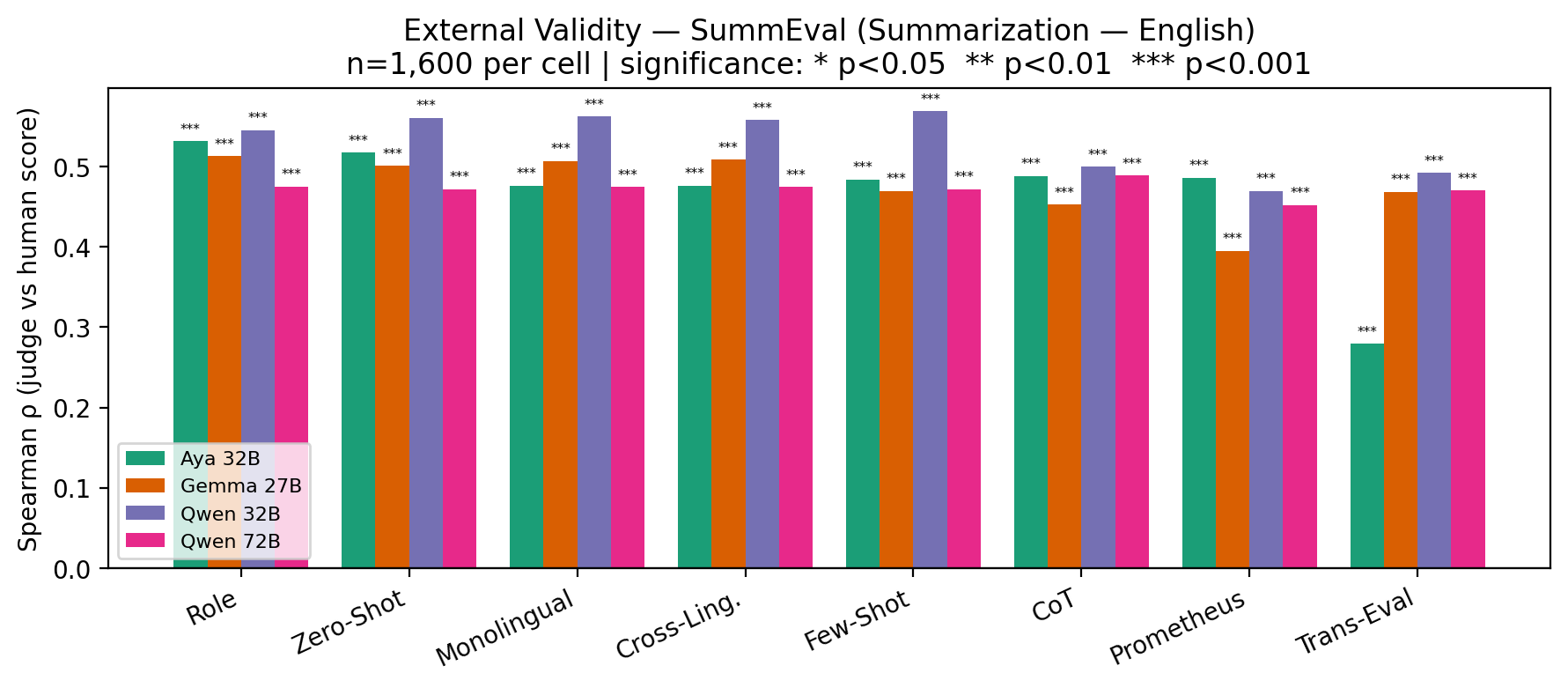}
  \caption{\textbf{External validity: SummEval
  (Summarization, English, $n=1{,}600$).}
  Per-example Spearman $\rho$ between judge scores and
  human quality annotations per strategy and judge model.
  Significance markers: * $p<0.05$, ** $p<0.01$,
  *** $p<0.001$.}
  \label{fig:validity_summeval}
\end{figure*}

\begin{figure*}[!t]
  \centering
  \includegraphics[width=0.9\textwidth]{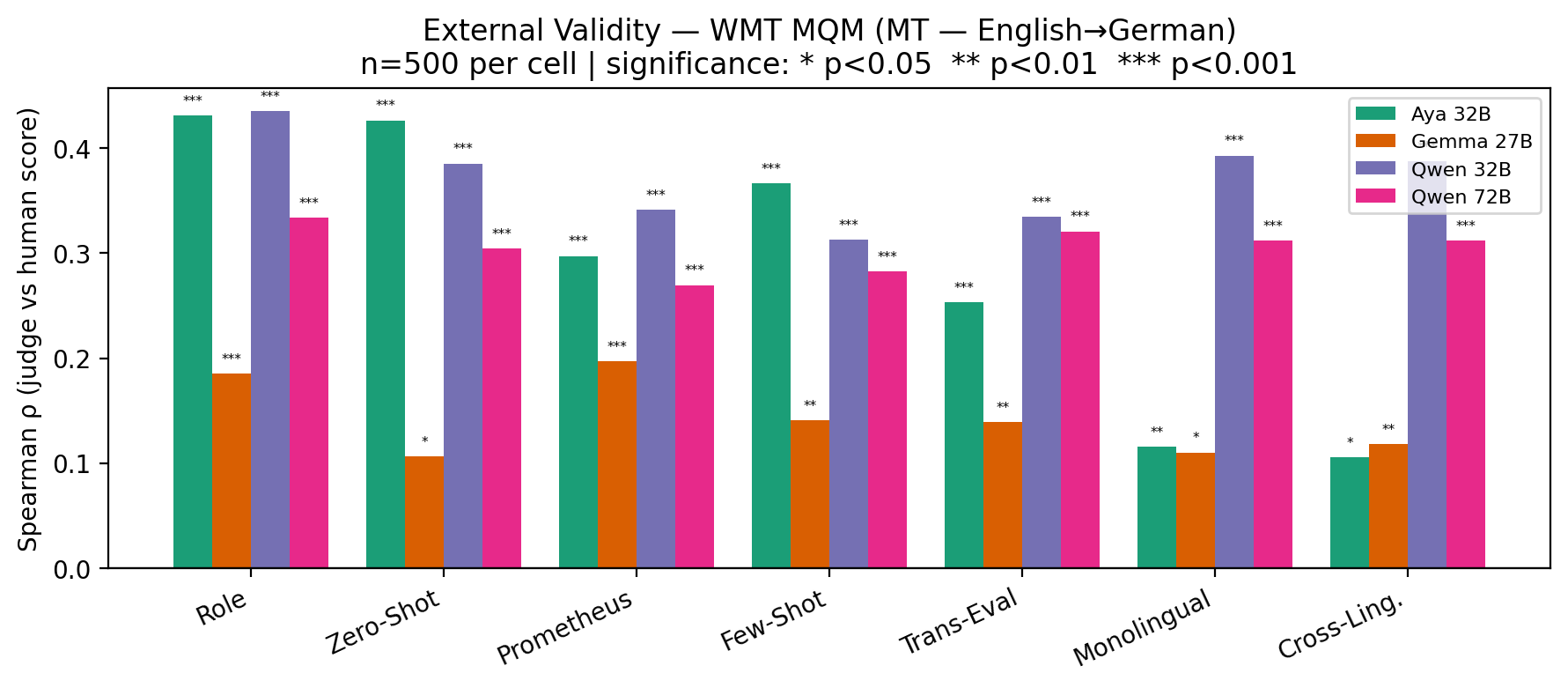}
  \caption{\textbf{External validity: WMT MQM
  (MT, English$\to$German, $n=500$).}
  Per-example Spearman $\rho$ between judge scores and
  expert MQM error annotations per strategy and judge model.}
  \label{fig:validity_wmt}
\end{figure*}

Figures~\ref{fig:validity_summeval} and~\ref{fig:validity_wmt}
show the per-example Spearman correlations underlying
Table~\ref{tab:validity} in the main text, split by dataset. Both
datasets are English-only, so results here complement, rather than
replace, the multilingual validation against SSA-MTE
(Section~\ref{sec:validity}). Correlations are broadly consistent
across strategies and judges in both datasets, with no strategy
performing catastrophically on either external source, suggesting the
patterns we observe are not artefacts of a single validation set.

\newpage
\section{Use of generative AI}

Generative AI tools were used at several stages during the preparation of this study. GitHub Copilot and Claude Code were used as coding assistants during the development of the evaluation pipeline and for debugging. Additionally, LLMs (Grammarly and Claude) were used to support the writing process, specifically for correcting spelling and grammatical inconsistencies, improving clarity, and reducing selected sections to comply with space constraints. Furthermore, all aspects of the research design, methodology, analysis, and conclusions presented in this study are the independent work of the author.

\end{document}